\documentclass[a4paper,10pt]{IEEEtran}

\usepackage[utf8]{inputenc} 
\usepackage[T1]{fontenc}
\usepackage{url}
\usepackage{bm}
\usepackage{ifthen}
\usepackage{cite}
\usepackage{hyperref}
\usepackage{amssymb}
\usepackage[cmex10]{amsmath} 
\newtheorem{lemma}{Lemma}

\newtheorem{theorem}{Theorem}

 \usepackage{algorithm}
 \usepackage{algpseudocode}
\makeatletter
\newcommand{\removelatexerror}{\let\@latex@error\@gobble}
\makeatother
\algnewcommand\algorithmicinput{\textbf{Input:}}
\algnewcommand\INPUT{\item[\algorithmicinput]}
\algnewcommand\algorithmicoutput{\textbf{Output:}}
\algnewcommand\OUTPUT{\item[\algorithmicoutput]}

\usepackage[]{graphicx}
\usepackage[]{color} 

\begin{document}
\title{Convergence Acceleration via Chebyshev Step:  \\ Plausible Interpretation of Deep-Unfolded Gradient Descent
}

\author{%
  \IEEEauthorblockN{
  		Satoshi Takabe\IEEEauthorrefmark{1}\IEEEauthorrefmark{2},
  		and Tadashi Wadayama\IEEEauthorrefmark{1}}\\
  \IEEEauthorblockA{\IEEEauthorrefmark{1}%
		Nagoya Institute of Technology,
		Gokiso, Nagoya, Aichi 466-8555, Japan,\\
 		\{s\_takabe, wadayama\}@nitech.ac.jp} \\
  \IEEEauthorblockA{\IEEEauthorrefmark{2}%
  		RIKEN Center for Advanced Intelligence Project,
  		Nihonbashi, Chuo-ku, Tokyo 103-0027, Japan
                }
}

\maketitle

\begin{abstract}
Deep unfolding is a promising deep-learning technique, whose network architecture is based on expanding the recursive structure of {existing iterative algorithms.} 
Although convergence acceleration is a remarkable advantage of deep unfolding, its theoretical aspects have not been revealed yet.
The first half of this study details the theoretical analysis of the convergence acceleration in deep-unfolded gradient descent (DUGD) whose trainable parameters are step sizes.
We propose a plausible interpretation of the learned step-size parameters in DUGD by introducing the principle of Chebyshev steps derived from Chebyshev polynomials.  
The use of Chebyshev steps in gradient descent (GD) enables us to bound the spectral radius 
of a matrix governing the convergence speed of GD, leading to a tight upper bound on the convergence rate.
The convergence rate of GD using Chebyshev steps is shown to be asymptotically 
optimal, although it has no momentum terms. 
We also show that Chebyshev steps numerically explain
 the learned step-size parameters in DUGD well.
In the second half of the study, 
 Chebyshev-periodical successive over-relaxation (Chebyshev-PSOR) is proposed 
for accelerating linear/nonlinear fixed-point iterations.
Theoretical analysis 
{and numerical} 
experiments indicate that Chebyshev-PSOR exhibits significantly faster convergence for various examples such as Jacobi method and proximal gradient methods. 
\end{abstract}

\section{Introduction}\label{sec:intro}

 Deep learning is emerging as a fundamental technology 
 in numerous fields such as image recognition, speech recognition, and natural language processing.
Recently, deep learning techniques have been applied to 
general computational processes that are not limited to neural networks.
The approach is called {\em differential programming}.
If the internal processes of an algorithm are differentiable, then
standard deep learning techniques
such as back propagation and stochastic gradient decent (SGD)
can be employed 
to adjust learnable parameters of the algorithm.
Data-driven tuning often improves the performance of the original algorithm, and
it provides flexibility to learn the environment in which the algorithm is operating.

{A set of standard deep learning techniques}
can be naturally applied to internal parameter optimization of 
a differentiable {\em iterative} algorithm such as iterative minimization algorithm.
By embedding the learnable parameters in an iterative algorithm, 
we can construct a flexible derived algorithm in a data-driven manner.
This approach is often called {\em deep unfolding}~\cite{gregor2010learning,hershey2014deep}.
In~\cite{gregor2010learning}, a deep-unfolded sparse signal recovery algorithm is proposed based on the iterative shrinkage-thresholding algorithm (ISTA); the proposed algorithm shows remarkable performance compared with the original ISTA.

Recently, numerous studies on signal processing algorithms focusing deep unfolding have 
been presented 
for sparse signal recovery ~\cite{sprechmann2015learning,xin2016maximal,borgerding2017amp,pmlr-v97-wu19b,ito2019trainable,takabe2020complex},
image recovery~\cite{shi2017deep,jin2017deep,mardani2018neural,kellman2019physics},
 wireless communications~\cite{nachmani2016learning,samuel2017deep,he2018model,8759948,yao2019sure,DBLP:conf/isit/WadayamaT19,takabe2020deep}, {and distributed computing~\cite{9157886}}.
Theoretical aspects of deep unfolding have also been investigated~\cite{chen2018theoretical,liu2018alista,mardani2019degrees}, which mainly focus on analyses of deep-unfolded sparse signal recovery algorithms.
More details on related works can be found 
in the excellent surveys on
deep unfolding~\cite{9020494, monga2019algorithm}.

Ito et al. proposed {\em trainable ISTA (TISTA)}~\cite{ito2019trainable} for sparse signal recovery which is an instance of deep-unfolded algorithms. This algorithm can be regarded as a proximal gradient method with trainable step-size parameters.
They showed that the convergence speed of signal recovery processes 
can be significantly accelerated with deep unfolding.
This phenomenon, called {\em convergence acceleration}, is the most significant
advantage of deep unfolding.

\begin{figure}[t]
   \centering
   \includegraphics[width=0.88\hsize]{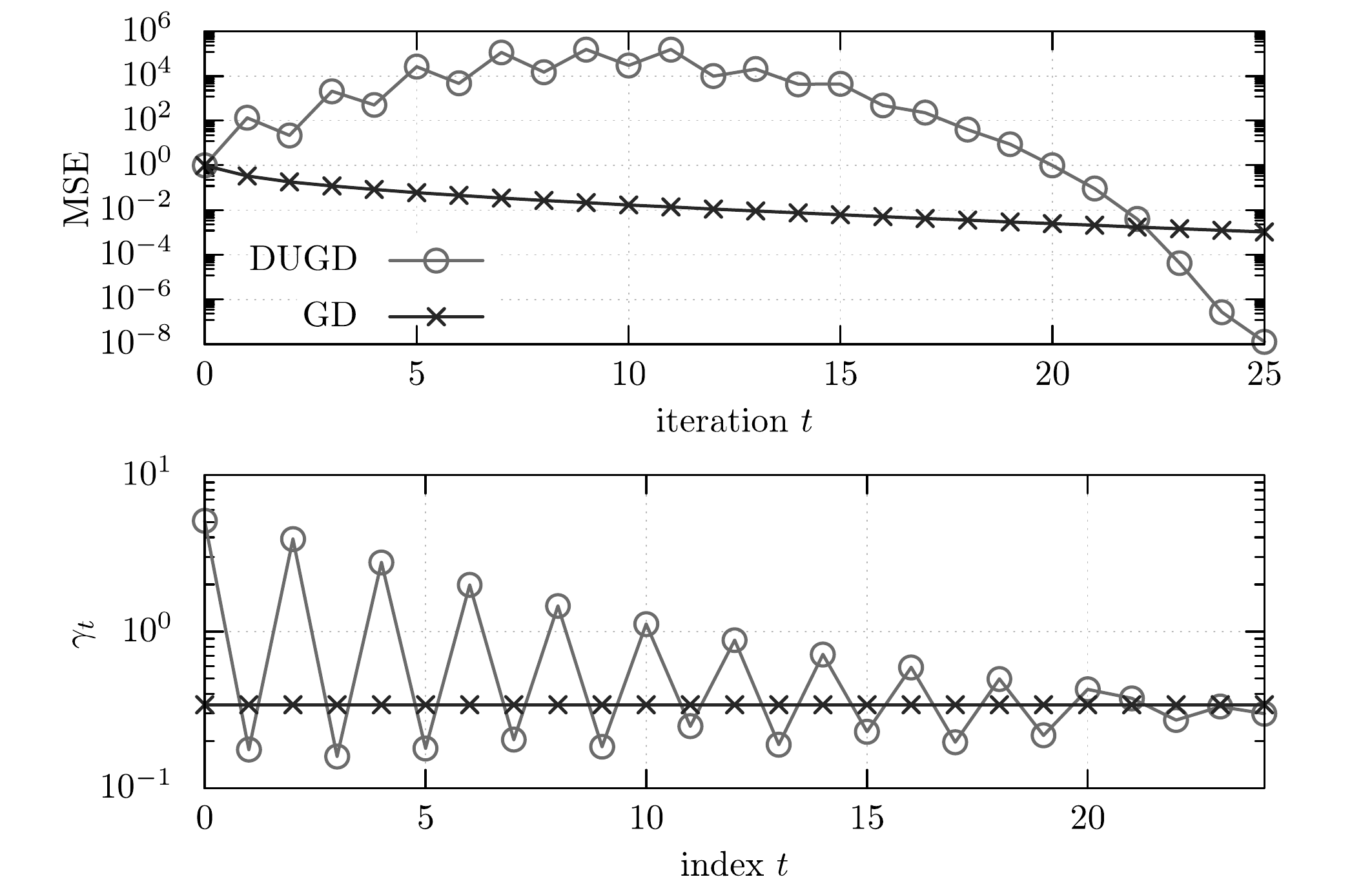}
    \caption{Mean squared error (MSE) (upper) and learned step-size parameters $\gamma_t$
    (lower) of deep-unfolded gradient descent (DUGD)
    and naive gradient descent (GD) with a constant step size.
    The objective function is a quadratic function with 300 variables. 
	{See Sec.~\ref{sec_predugd} for details.}
}
    \label{fig_ga}
\end{figure}

However, a clear explanation on {\em how convergence acceleration emerges when
using deep unfolding is still lacking}. 
When plotting the learned step sizes of TISTA as a function of the iterative step, we often 
obtain a zig-zag shape (Fig.11 of~\cite{ito2019trainable}), i.e., non-constant step sizes. 
{Another} example of convergence acceleration is shown in Fig.\ref{fig_ga}. 
We can observe that the DUGD provides much faster convergence to the optimal point 
compared with a naive GD with the optimal constant step size. A zig-zag shape of a learned 
step-size sequence can be seen in Fig.\ref{fig_ga} (lower).
The zig-zag strategy on step sizes seems preferable to accelerate the convergence; however, 
a reasonable interpretation still {remains an open problem}.


The primary goal of this study is to provide a theoretically convincing interpretation of learned parameters and
to answer the question regarding the mechanism of the convergence acceleration.
We expect that deepening our understanding regarding convergence acceleration is of crucial importance 
because knowing the mechanism would lead to a novel way for improving 
convergence behaviors of numerous iterative signal processing algorithms.

The study is divided into two parts. In the first part, we will pursue 
our primary goal, i.e., to provide a plausible theoretical interpretation of learned step sizes. 
The main contributions of this part are 
{1) a special step size sequence called {\em Chebyshev steps} is introduced by extending an acceleration method for average consensus~\cite{8678811}, which can accelerate the convergence speed of GD without momentum terms and} 
{
2) a learned step size in DUGD can be well explained by Chebyshev steps,
which can be considered a promising answer to the question posed above. 
The Chebyshev step is naturally derived by minimizing an upper bound on the spectral radius {dominating the convergence rate using Chebyshev polynomials}.

The second part is devoted to our secondary goal, i.e., to extend the theory of Chebyshev steps.
We especially focus on accelerating the linear/nonlinear fixed-point iterations using Chebyshev steps. 
The main contributions of the second part are 
1) a novel method, called {\em Chebyshev-periodical successive over-relaxation (PSOR)}, 
for accelerating the convergence speed of 
fixed-point iterations and 2) its local convergence analysis are proposed.
Chebyshev-PSOR can be regarded as a variant of
SOR or Krasnosel'ski\v{\i}-Mann iteration~\cite{mann1953mean}.
One of the most notable features of Chebyshev-PSOR is that it can be applied to
nonlinear fixed-point iterations in addition to linear fixed-point iterations.
It appears particularly effective for accelerating proximal gradient methods 
such as ISTA.

This paper is organized as follows. 
In Section~\ref{sec_2}, we describe brief reviews of background materials as preliminaries. 
In Section~\ref{sec_dugd}, we provide a theoretical interpretation of learned step-size parameters in DUGD by introducing Chebyshev steps. 
In Section~\ref{sec_fpi}, we introduce Chebyshev-PSOR for fixed-point iterations as an application of Chebyshev steps. 
The last section provides a summary of this paper.

\section{Preliminaries}\label{sec_2}


\subsection{Notation}
In this paper, $ \bm x=(x_1,x_2,\dots,x_n)^T$ represents an $n$-dimensional column vector and 
$\bm  A=(a_{ij})^{1\le i\le m}_{1\le j\le n}$ denotes a matrix of size $m\times n$.
In particular, $\bm I_n$ denotes {the} identity matrix of {size} $n$. 
{For a complex matrix $\bm  A \in\mathbb{C}^{n\times n}$, $\bm  A^{\rm H}$ denotes a Hermitian transpose of $\bm  A$.}
{For $\bm A\in\mathbb{C}^{n\times n}$ with real eigenvalues,} 
we define $\lambda_{max}(\bm A)$ and $\lambda_{min}(\bm A)$ as the maximum and minimum eigenvalues of $\bm A$, respectively.
Then, the condition number of $\bm A$ is given as $\kappa(\bm A):= |\lambda_{max}(\bm A)/\lambda_{min}(\bm A)|$.
The spectral radius of $\bm A$ is given as 
$\rho(\bm A):= \max_{n=1,\dots, n}|\lambda_i|$, 
where $\lambda_i$ is an eigenvalue of the matrix $\bm A$.
For matrix $\bm A \in\mathbb{C}^{n\times n}$, $\|\bm A\| := \sup_{\|\bm v\|_2=1;\bm{v}\in\mathbb{C}^n} \|\bm A \bm v\|_2$ represents the operator norm (spectral norm) of $\bm A$. 
We define $\mathcal{N}(\mu,\sigma^2)$ as the Gaussian distribution with mean $\mu$ and variance $\sigma^2$.

\subsection{Deep unfolding}

In this subsection, we briefly introduce the principle of deep unfolding and demonstrate it using {a toy problem}.

\subsubsection{Basic idea}
Deep unfolding is a deep learning technique based on existing iterative algorithms.
Generally, iterative algorithms have some parameters 
that {control} convergence properties.
In deep unfolding, we first expand the recursive structure to the unfolded feed-forward network, in which we can embed some trainable parameters to increase the flexibility of {the iterative algorithm}.  
These embedded parameters can be trained by standard deep learning techniques 
if the algorithm consists of only differentiable sub-processes.
Figure~\ref{fig_dua} shows a schematic diagram of (a) an iterative algorithm with {a set of} parameters $\bm \Theta$ and (b) an example of deep-unfolded algorithm.
In the deep-unfolded algorithm, the embedded trainable parameters $\{\bm \Theta_t\}_{t=0}^{T-1}$ are updated to decrease 
the loss function between the true solution $\bm{\tilde x}$ and output $\bm{x_T}$ by an optimizer.
This training process is executed using supervised data $(\bm{\tilde x},\bm{\tilde y})$, where $\bm{\tilde y}$ is an input of the algorithm and
 $\bm{\tilde x}$ is the corresponding true solution.


 \begin{figure}[t]
   \centering
   \includegraphics[width=0.85\hsize]{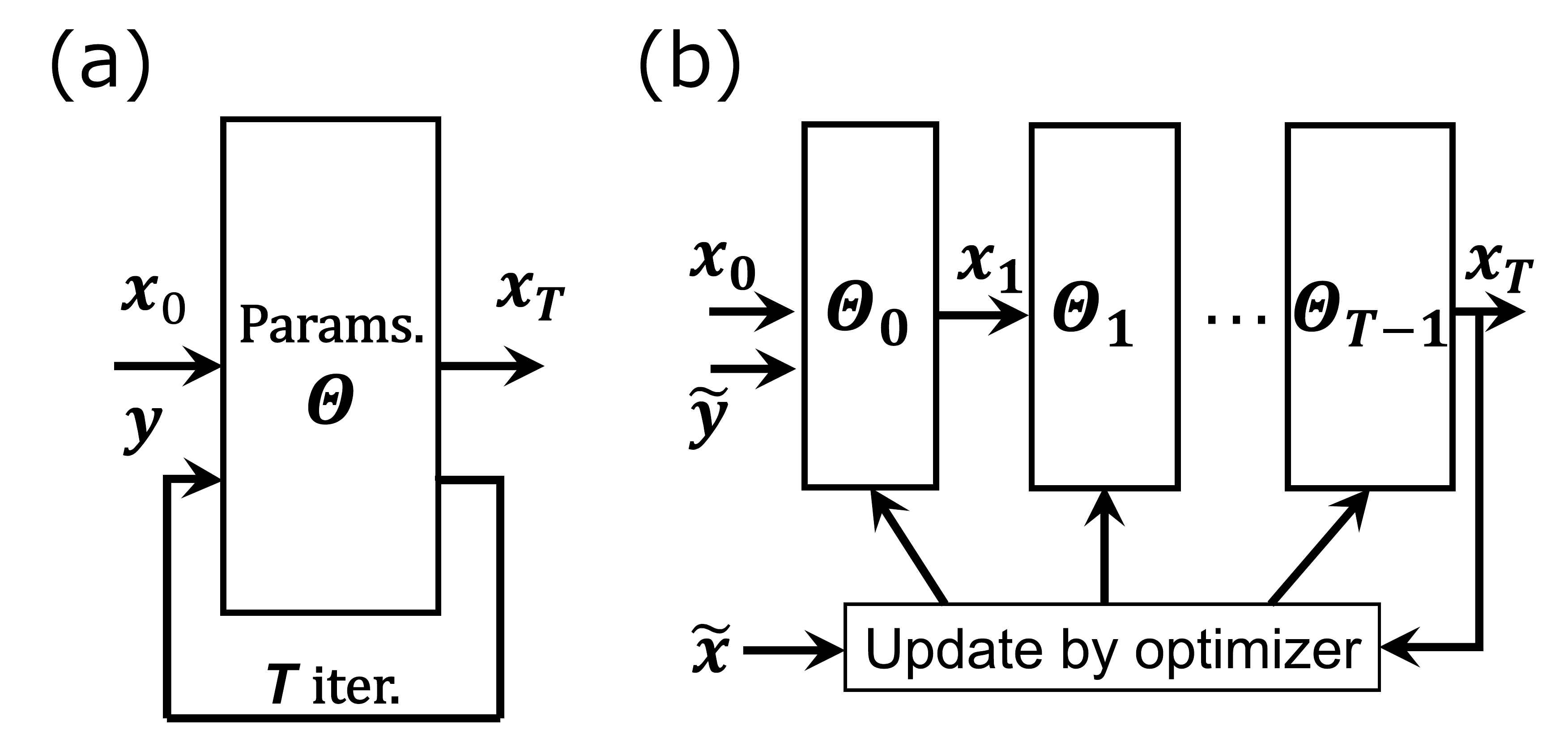}
    \caption{Schematic {diagrams} of (a) an iterative algorithm with parameters $\bm \Theta$ and (b) deep-unfolded algorithm with trainable parameters $\{\bm \Theta_t\}_{t=0}^{T-1}$. In the deep-unfolded algorithm, trainable parameters are updated to decrease 
the loss function between a true solution $\bm{\tilde x}$ and output $\bm x_T$. 
}
    \label{fig_dua}
    \vskip -0.2in
\end{figure}

\subsubsection{Example of deep unfolding}\label{sec_predugd}
We turn to an example of a deep-unfolded algorithm based on GD~\footnote{A sample code is available at \url{https://github.com/wadayama/Chebyshev}.}.
Let us consider a simple least mean square (LMS) problem written as 
\begin{equation}
\bm {\hat \beta}:=\arg \min_{\bm  \beta\in \mathbb{R}^n} \frac{1}{2}\| \bm y- \bm H  \bm \beta\|_2^2,
\label{eq_LMS}
\end{equation}
where $\bm  y=\bm  H  \bm \beta^\ast\in \mathbb{R}^m$ is a noiseless measurement vector with the measurement matrix $\bm {H}\in \mathbb{R}^{m\times n}$ ($m\ge n$) {and the original signal $\bm \beta^\ast\in\mathbb{R}^n$}.  
{ In the following experiment, $\bm  H$ is sampled from 
a Gaussian random matrix ensemble}, whose elements {independently follows $\mathcal N(0,1/n)$}

Although the solution of (\ref{eq_LMS}) is explicitly given by $\bm {\hat \beta}=(\bm  H^{T}\bm  H)^{-1} \bm H^{T} \bm y$,
{GD can be used to evaluate its solution.}
The recursive formula of GD is given by  
\begin{equation}
\bm {\beta}^{(t+1)}=  \bm \beta^{(t)}{+}\gamma  \bm H^{T}( \bm y -  \bm H  \bm \beta^{(t)}) \:(t=0,1,2,\dots),
\label{eq_GD1}
\end{equation}
where $\bm \beta^{(0)}$ is an initial vector, and $\gamma$ is a step-size parameter.

By applying the principle of deep unfolding, we introduce a deep-unfolded GD (DUGD) by 
\begin{equation}
\bm{\beta}^{(t+1)}= \bm \beta^{(t)}{+} \gamma_t  \bm H^{T}( \bm y -  \bm H  \bm \beta^{(t)}),
\label{eq_GD2}
\end{equation}
where $\gamma_t$ is a trainable step-size parameter that depends on the iteration index $t$.
The parameters $\{\gamma_t\}_{t=0}^{T-1}$ can be trained by minimizing the mean squared error (MSE) $\|\bm {\tilde \beta}-\bm \beta^{(T)}\|_2^2/(Bn)$ between the estimate $\bm {\beta}^{(T)}$ and $\bm {\tilde \beta}$ 
in a mini batch containing $B$ supervised data $\{(\bm {\tilde{\beta}}, \bm {\tilde{y}})\}$.  
Especially, we employ \emph{incremental training}~\cite{ito2019trainable} in the training process {to avoid a gradient vanishing problem and escape from local minima of the loss function}.
During incremental training, the number of trainable parameters increases at every $B$ mini batch; that is, the DUGD of $T$ iterations (or layers) is trained from the $B(T-1)$th mini batch to the $(BT-1)$th mini batch, which we call the $T$th generation of incremental training.  
After the $T$th generation ends, we add an initialized parameter $\gamma_{T}$ to the learned parameters $\{\gamma_{t}\}_{t=0}^{T-1}$ to start the next generation.

 Figure~\ref{fig_ga} shows empirical results of the MSE performance (upper) and learned parameters $\{\gamma_t\}_{t=0}^{T-1}$ (lower) of DUGD ($T\!=\!25$) and the naive GD with optimal constant step size
$\gamma \!=\! 2/(\lambda_{max}(\bm  H^{T} \bm H)+\lambda_{min}(\bm  H^{T} \bm H))$ 
  when $(m,n)\!=\!(300,600)$.
In the training process, the initial values of $\{\gamma_t\}$ were set to $0.3$, and 
$500$ mini batches of size $200$ and Adam optimizer~\cite{kingma2014adam} with learning rate $0.002$ were used in each generation.
We found that the learned parameter sequence had a \emph{zig-zag shape}
{and that DUGD yields a much steeper error curve after 20-25 iterations 
compared with the naive GD with a constant step size.}
These learned parameters are not intuitive or interpretable from conventional theory.
{Unveiling the reason for the zig-zag shape is the main objective of the next section.}

%

\subsection{Brief review of accelerated gradient descent}

{In the first part of this study, we will focus on GD for minimizing 
a quadratic objective function
$\bm x^{\rm H}  \bm A \bm x/2$, 
where {$\bm  x \in\mathbb C^{n}$} and $\bm A \in\mathbb C^{n\times n}$ is a Hermitian matrix.}
{Here, we will review first-order methods that use only first-order gradient information in their update rules.}

\subsubsection{GD with optimal constant step}

{An iterative process} of a naive GD with the constant step size $\gamma$ is given by 
\begin{equation}
	\bm {x}^{(t+1)} = (\bm {I}_n - \gamma \bm {A}) \bm {x}^{(t)} := \bm W\bm x^{(t)}.
	\label{eq_gd1}
\end{equation}
It is known~\cite[Section 1.3]{9781886529052} that the convergence rate $R_\mathrm{s}$ with 
an optimal step size $\gamma=\gamma_s$ is given as
\begin{equation}
	R_{\mathrm s} = \frac{\kappa-1}{\kappa+1},\quad \gamma_{\mathrm s} := \frac{2}{\lambda_{max}(\bm A)+\lambda_{min}(\bm A)}.
	\label{eq_gd2}
\end{equation}
where $\kappa := \kappa(\bm A)$. 

\subsubsection{Momentum method}

{One of the most well-known variants of GD is 
the {\em momentum method}~\cite{1986Natur.323..533R} (or heavy ball method~\cite{polyak1964some}), which is given as}
\begin{equation}
	\bm {x}^{(t+1)} = (\bm {I}_n - \gamma' {A}) \bm {x}^{(t)} +\beta(\bm {x}^{(t)}-\bm {x}^{(t-1)}),
	\label{eq_mom}
\end{equation}
where $\bm  x^{(-1)}= 0$, $\gamma':=4/(\sqrt{\lambda_{min}(\bm A)}+\sqrt{\lambda_{max}(\bm A)})^2$, and 
$\beta:=((\sqrt{\kappa}-1)/(\sqrt{\kappa}+1))^2$~\cite{polyak1964some}. 
{The momentum method exhibits faster convergence than the naive GD because of the \emph{momentum term} $\bm {x}^{(t)}-\bm {x}^{(t-1)}$}.
{It was proved that the momentum method achieves a convergence rate
 $R_{*} = {(\sqrt\kappa-1)}/{(\sqrt\kappa+1)}$.
which coincides with the lower bound of the convergence rate of 
the first-order methods~\cite{nesterov1998introductory}. This implies that the momentum method 
is an optimal first-order method in terms of the convergence rate.}

\subsubsection{Chebyshev semi-iterative method}

{The Chebyshev semi-iterative method is another first-order method with
an optimal rate which is defined by}
\begin{align}
	\bm {x}^{(t+1)} &= (\bm {I}_n - \gamma'_{t+1} \bm {A}) \bm {x}^{(t)} +(\gamma'_{t+1}-1) (\bm {x}^{(t)}-\bm {x}^{(t-1)}),\nonumber\\
	\gamma'_{t+1} &= \frac{4}{4-\xi^2\gamma'_t}\: (t\ge 2), \label{eq_chsemi}
\end{align}
where {the initial conditions are given by} $ \bm x^{(-1)}= 0$, $\gamma'_1=1$,  $\gamma'_2=2/(2-\xi^2)$, and $\xi = 1-1/\kappa$~\cite{Golub1989}.
The Chebyshev semi-iterative method also achieves the convergence rate $R_*$.
\section{Chebyshev steps for gradient descent}\label{sec_dugd}

In this section, we introduce Chebyshev steps that accelerate the convergence speed of GD and {discuss the relation between Chebyshev steps and
learned step-size sequences in DUGD.}

\subsection{Deep-unfolded gradient descent}\label{sec_dugdpre}
We consider the minimization of a convex quadratic function 
$f(\bm {x}) := \bm {x}^{\rm H} \bm {A} \bm {x}/2$ where $ \bm A \in \mathbb{C}^{n \times n}$ is a Hermitian positive definite matrix and $\bm {x}^{\ast} = \bm {0}$ is its solution. {Note that this minimization problem corresponds to the LMS problem (\ref{eq_LMS}) under proper transformation.}

 GD with the step-size sequence $\{\gamma_t\}$ is given by 
\begin{equation}
\bm {x}^{(t+1)} = (\bm {I}_n - \gamma_t\bm  {A}) \bm {x}^{(t)} :=  \bm W^{(t)}\bm {x}^{(t)},\label{eq_W}
\end{equation}
where $\bm {x}^{(0)}$ is an arbitrary point in $\mathbb{C}^n$.

In DUGD, we first fix the total number of iterations $T(\ll n)$~\footnote{If $n<T$, GD always converges to the optimal solution {after $n$ iterations by setting} the step sizes to the reciprocal of eigenvalues of $ \bm A$. We thus omit this case.} and train the step-size parameters $\{\gamma_t\}_{t=0}^{T-1}$.
Training these parameters is typically executed by minimizing the MSE loss function 
{$L( \bm x^{(T)}):=\| \bm x^{(T)}-\bm {x}^{\ast} \|_2^2/n$} between the output $ \bm x^{(T)}$ of DUGD and the true solution $\bm {x}^{\ast} = \bm {0}$.
{To ensure convergence of DUGD, 
we assume {\em periodical trainable step-size parameters},
i.e, $\gamma_{t'} = \gamma_{t}$ $(t \in \{0, 1, \ldots, T-1\}, t'= 0, 1, \ldots)$ 
where $t'\equiv t $ (mod $T$).
This means that we enforce the equal constraint $\gamma_{t'} = \gamma_{t}$ in training 
processes and that
DUGD uses a learned parameter sequence $\{\gamma_t\}_{t=0}^{T-1}$ repeatedly for $t>T$. }
{A similar idea is presented in~\cite{8678811,9157886} for average consensus.}
In this case, the {output} after every $T$ step is given as
\begin{equation}
	\bm {x}^{((k+1)T)} = \left(\prod_{t=0}^{T-1} \bm W^{(t)}\right)\bm {x}^{(kT)} :=  \bm Q^{(T)}\bm {x}^{(kT)},
\label{eq_TPGT}
\end{equation}
for any $k=0,1,2,\dots$. Note that $\bm {Q}^{(T)}$ is a function of step-size parameters $\{\gamma_t\}_{t=0}^{T-1}$.
We aim to show that a proper step-size sequence $\{\gamma_t\}_{t=0}^{T-1}$ accelerates the convergence speed, i.e., reduces the spectral radius $\rho(\bm Q^{(T)})$.  

\subsection{Chebyshev steps}\label{sec_chst}
 {In this subsection, our aim is to determine a step-size sequence that reduces the spectral radius $\rho( \bm Q^{(T)})$ to understand the nontrivial step-size sequence of DUGD.
In the simplest $T=1$ case, the optimal step size is analytically given by {$\gamma_t\!=\!\gamma_{\mathrm s}$ in (\ref{eq_gd2})}.
When $T\ge 2$, however, minimizing $\rho( \bm Q^{(T)})$ with respect to $\{\gamma_t\}_{t=0}^{T-1}$ becomes a non-convex problem in general. We thus cannot rely on numerical optimization methods. 
We alternatively introduce a step-size sequence that tightly bounds the spectral radius from above {and analyze its convergence rate}.
{These ideas follow the worst-case optimal (WO) interpolation for average consensus~\cite{8678811}. 
Whereas the WO interpolation is considered only for a graph Laplacian, the analysis in this paper treats GD with a general Hermitian matrix whose minimum eigenvalue may not be equal to zero.}

Recall that the Hermitian positive definite matrix $ \bm A$ has $n$ positive eigenvalues
 $(0<)\lambda_1 = \lambda_{min}(\bm A)\le \lambda_2
 \le \dots\le \lambda_n= \lambda_{max}(\bm A)$ including degeneracy.
Hereafter, we assume that $\lambda_1\neq\lambda_n$ to avoid a trivial case.

{To calculate eigenvalues of a matrix function, we use the following lemma~\cite[p. 4-11, 39]{hogben2013handbook}. 
\begin{lemma}\label{lem_a}
For a polynomial $p(x)$ with complex coefficients, if $\lambda$ is an eigenvalue of $ \bm A$ associated with eigenvector $\bm  x$, then $p(\lambda)$ is an eigenvalue of the matrix $p(\bm A)$ associated with the eigenvector $\bm x$.
\end{lemma}
Then, we have}
\begin{equation}
\rho(\bm  Q^{(T)})=\max_{i=1,\dots,n}\left|\prod_{t=0}^{T-1}(1-\gamma_t\lambda_i)\right|
=\max_{i=1,\dots,n}|\beta_T(\lambda_i)|, \label{eq_rho}
\end{equation}
where $\beta_T(\lambda):= \prod_{t=0}^{T-1}(1-\gamma_t\lambda)$ is a function containing $T$ step-size parameters $\{\gamma_t\}_{t=0}^{T-1}$.

{In the general case in which $T\ge 2$,
we focus on the step-size sequence which minimizes the upper bound $\rho^{\mathrm{upp}}( \bm Q^{(T)})$ of the spectral radius $\rho( \bm Q^{(T)})$. The upper bound is} given by 
\begin{align}
\rho( \bm Q^{(T)})&=\max_{i=1,\dots,n}\left|\beta_T(\lambda_i)\right|\nonumber\\
&\le  \max_{ \lambda\in [\lambda_1,\lambda_n] }|\beta_T(\lambda)|:= \rho^{\mathrm{upp}}( \bm Q^{(T)}).\label{eq_upp}
\end{align}

{In the following discussions, we use Chebyshev polynomials  $C_n(x)$~\cite{Mason03}  (see App.~\ref{sec_poly} in Supplemental Material).
Chebyshev polynomials have the minimax property, in which $2^{1-T}C_n(x)$ has the smallest $\ell_\infty$-norm in the Banach space $B[-1,1]$ (see Thm.~\ref{thm_ch}).}
Using the affine transformation $z = (\lambda_n-\lambda_1)x/2+(\lambda_n+\lambda_1)/2$, the monic polynomial is given as
\begin{equation}
\phi(z) := 2^{1-T}C_T\left(\frac{2z-\lambda_n-\lambda_1}{\lambda_n-\lambda_1}\right),
\label{eq_che_phi}
\end{equation} 
has the smallest $\ell_\infty$-norm in $B[\lambda_1,\lambda_n]$.
Let {$\{\gamma_t^{ch}\}_{t=0}^{T-1}$} be reciprocals to the zeros of $\phi(z)$, i.e., $\phi(z)=\prod_{t=0}^{T-1}[z-(\gamma_t^{ch})^{-1}]$.
Then, from (\ref{eq_ch3}), $\{\gamma_t^{ch}\}$ are given by 
\begin{equation}
\gamma_t ^{ch}= \left[\frac{\lambda_n+\lambda_1}{2}+\frac{\lambda_n-\lambda_1}{2}\cos\left(\frac{2t+1}{2T}\pi\right)\right]^{-1}.
\label{eq_chstep0}
\end{equation}
In this study, this sequence is called \emph{Chebyshev steps}. 
{Note that this sequence is also proposed for accelerated Gibbs sampling~\cite{Fox2013}.} 

As described above, Chebyshev steps minimize the $\ell_\infty$-norm in $B[\lambda_1,\lambda_n]$ in which a function is represented by $\phi(z)$.
Although it is nontrivial that Chebyshev steps also minimize $\rho^{\mathrm{upp}}(\bm Q^{(T)})$, we can prove that this is true.

\begin{theorem}\label{thm_step}
For a given $T\in\mathbb{N}$, we define \emph{Chebyshev steps} 
$\{\gamma_t^{ch}\}_{t=0}^{T-1}$ \emph{of length $T$} as
\begin{equation}\label{eq_chstep}
\gamma_t^{ch}
    :=\left[ \lambda_+(\bm A) + \lambda_- (\bm A)\cos \left(\frac{2t+1}{2T} \pi \right) \right]^{-1},
\end{equation}
where
\begin{equation}
\lambda_\pm(\bm A) :=\frac{1}{2} ({\lambda_{max}(\bm A) \pm \lambda_{min}(\bm A)}).\label{eq_lam1} \\
\end{equation}
Then, the Chebyshev steps form a sequence that minimizes the upper bound $\rho^{\mathrm{upp}}( \bm Q^{(T)})$ of the spectral radius of $\bm  Q^{(T)}$.
\end{theorem}

The proof is shown in App.~\ref{app_1} {in Supplemental Material}.
Note that the Chebyshev step {is identical} to the optimal constant step size {$\gamma_t=\gamma_{\mathrm s}$ (\ref{eq_gd2})} when $T=1$.
Chebyshev steps are a natural extension of the optimal constant step size.

\begin{figure}[t]
   \centering
   \includegraphics[width=0.88\hsize]{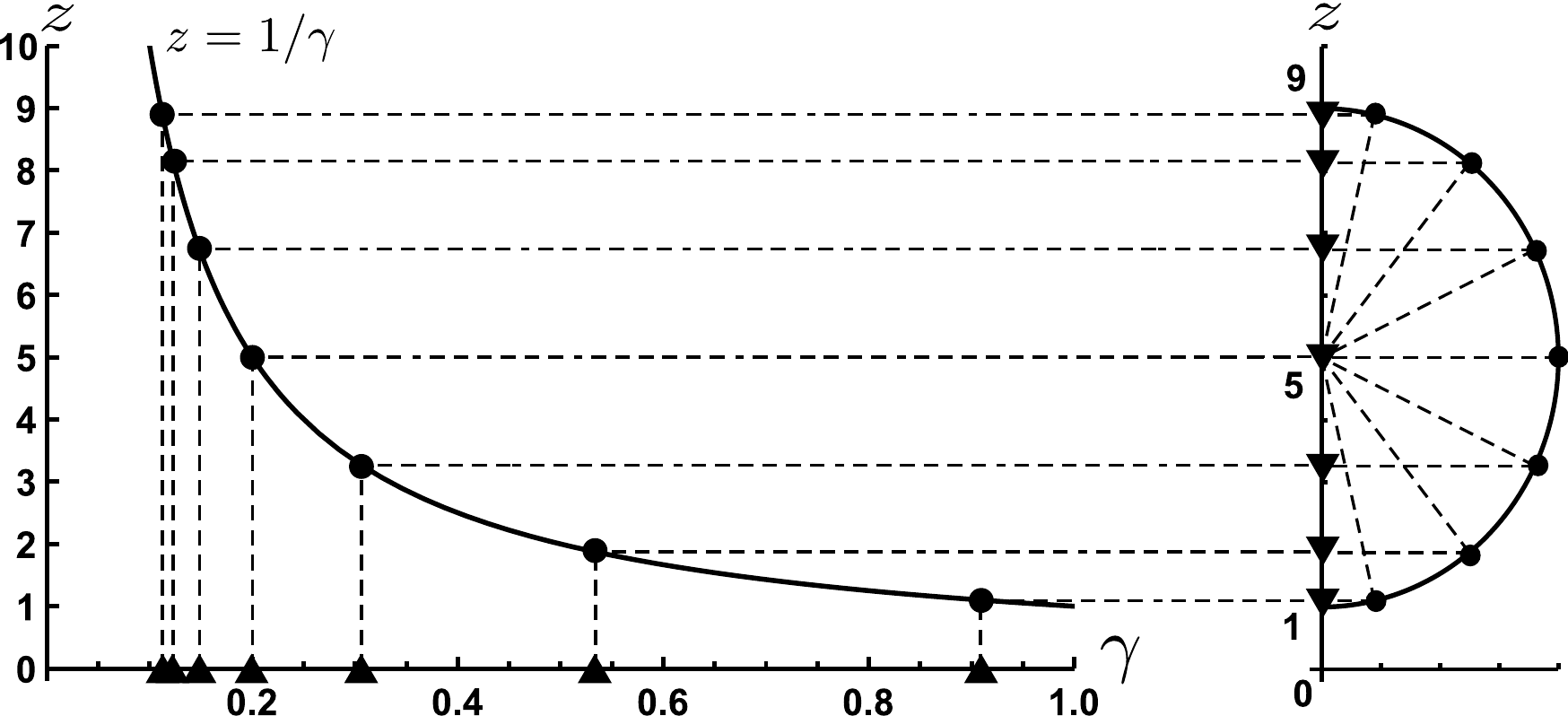}
    \caption{Chebyshev nodes $\{z_t^{ch}\}_{t=0}^{T-1}$ (right; lower triangle) and corresponding Chebyshev steps $\{\gamma_t^{ch}\}_{t=0}^{T-1}$ (left; upper triangle) when $T=7$, $\lambda_1=1$, and $\lambda_n=9$. 
}
    \label{fig_poi}
\end{figure}

The reciprocal of the Chebyshev steps $z_t^{ch}:=(\gamma_t^{ch})^{-1}$ corresponds to Chebyshev nodes as zeros of $\phi(z)$.
Figure~\ref{fig_poi} shows the Chebyshev nodes and Chebyshev steps when $T=7$, $\lambda_1=1$, and $\lambda_n=9$.
A Chebyshev node is defined as a point which is projected onto an axis from a point of degree $\theta_t=(2t+1)\pi/(2T)$ on a semi-circle (see right part of Fig.~\ref{fig_poi}). 
The Chebyshev nodes are located symmetrically with respect to the center of the semi-circle corresponding to $(\gamma_{\mathrm s})^{-1}=(\lambda_1+\lambda_n)/2$.
The Chebyshev steps are given by $\gamma_t^{ch}=(z_t^{ch})^{-1}$, which is shown in the left part of Fig.~\ref{fig_poi}.
We found that the Chebyshev steps are widely located 
compared with the optimal constant step size $\gamma_{\mathrm s}=1/5$.

\subsection{Convergence analysis}\label{sec_ca}

We analyze the convergence rate of  GD with Chebyshev steps (CHGD).  
{Because $\bm A$ is {a Hermitian matrix}, $\bm Q^{(T)}$ is a normal matrix and $\| \bm Q^{(T)}\|_2=\rho( \bm Q^{(T)})$~\cite[p. 8-7, 19]{hogben2013handbook} holds.}
Let $ \bm Q^{(T)}_{\mathrm{ch}}$ be the matrix $\bm {Q}^{(T)}$ {including the Chebyshev steps of length $T$.}
Then, we have
\begin{align}
\| \bm x^{( (k+1)T)}-\bm {x}^{\ast} \|_2 
&= \left\| \bm Q^{(T)}_{\mathrm{ch}}  \bm x^{( kT)} \right\|_2 \nonumber\\
&\le \left\| \bm Q^{(T)}_{\mathrm{ch}} \right\| \| \bm x^{( kT)} \|_2 \nonumber\\
&= \rho( \bm Q^{(T)}_{\mathrm{ch}})  \| \bm x^{( kT)} \|_2 \nonumber\\
&\le \rho^{\mathrm{upp}}( \bm Q^{(T)}_{\mathrm{ch}}) \| \bm x^{( kT)} -\bm {x}^{\ast} \|_2,\label{eq_dumse}
\end{align}
{where the second equality is due to {$\| \bm Q^{(T)}_{\mathrm{ch}}\|_2=\rho( \bm Q^{(T)}_{\mathrm{ch}})$.} 
This indicates that the MSE of every $T$ iteration is upper bounded by $\rho^{\mathrm{upp}}(\bm  Q^{(T)}_{\mathrm{ch}})$.

{Calculating $\rho^{\mathrm{upp}}(\bm  Q^{(T)}_{\mathrm{ch}})$ is similar to that in~\cite{8678811}.  Finally, we get the following lemma.}
\begin{lemma}\label{lem_qt}
For $T\ge 1$ and $\lambda\in[\lambda_1,\lambda_n]$, we have
\begin{equation}
|\beta_T^{ch}(\lambda)|\le \rho^{\mathrm{upp}}(\bm Q^{(T)}_{\mathrm{ch}}) 
=\mathrm {sech}\left[T \cosh^{-1}\left(\frac{\kappa+1}{\kappa-1}\right) \right] .
\end{equation} 
\end{lemma}
{See App.~\ref{app_rate} in Supplemental Material for the proof. }

 \begin{figure}[t]
   \centering
   \includegraphics[width=0.88\hsize]{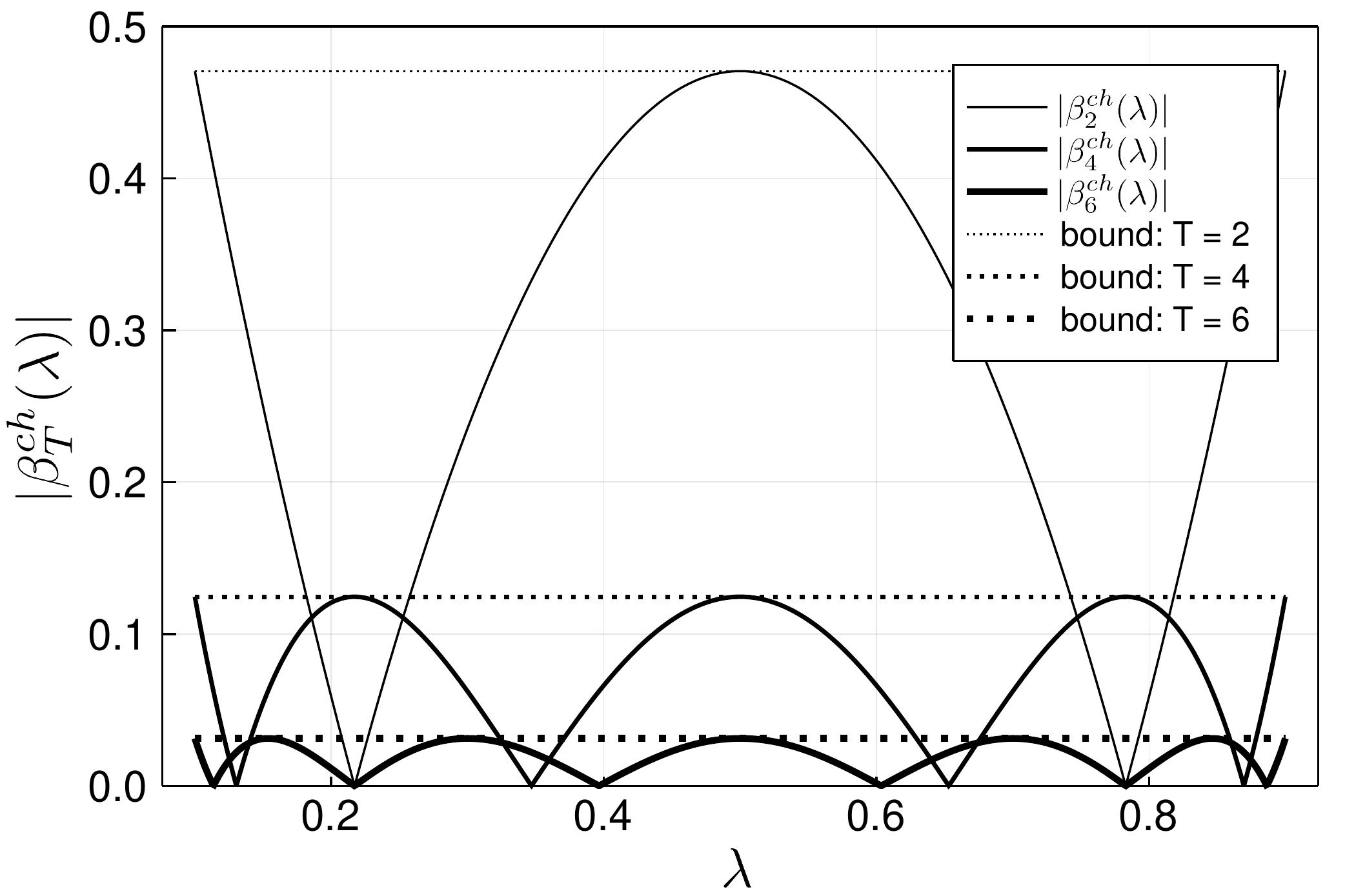}
    \caption{ The absolute value of $\beta_T^{ch}(\lambda)$ $(T = 2, 4, 6)$ where $\lambda_{min}(\bm A) = 0.1$
	and $\lambda_{max}(\bm A) = 0.9$. The dashed lines represent the upper bound $\rho^{\mathrm{upp}}( \bm Q^{(T)}_{\mathrm{ch}})$ for $T = 2, 4, 6$.
}
    \label{fig_bet}
    \vskip -0.2in
\end{figure}

{Let $\beta^{ch}_T(\lambda)$ be the function $\beta_T(\lambda)$ with Chebyshev steps.}
Figure~\ref{fig_bet} shows $|\beta_T^{ch}(\lambda)|$ and its upper bound $\rho^{\mathrm{upp}}( \bm Q^{(T)}_{\mathrm{ch}})$ for $T=2,4,6$ when $\lambda_{min}(\bm A) = 0.1$ and $\lambda_{max}(\bm A) = 0.9$. 
We find that $|\beta_T^{ch}(\lambda)|$ reaches its maximum value $\rho^{\mathrm{upp}}( \bm Q^{(T)}_{\mathrm{ch}})$ several times due to the extremal point property of Chebyshev polynomials.
Moreover, because {$\mathrm{sech}(x)< 1$ holds for $x>0$}, $\rho^{\mathrm{upp}}( \bm Q^{(T)}_{\mathrm{ch}})<1$ holds.
This indicates that CHGD always converges to an optimal point.

We first compare {the convergence rate obtained in the above argument} with the spectral radius of the optimal constant step size. 
Letting $ \bm Q^{(T)}_{\mathrm{const}}$ be $ \bm Q^{(T)}$ with the optimal constant step size $\gamma_0\!=\!\dots\!=\!\gamma_{T-1}\!=\!\gamma_{\mathrm s}$, we get $\rho( \bm Q^{(T)}_{\mathrm{const}}) = [{(\kappa-1)}/{(\kappa+1)]^T}.$
Then, we can prove the following theorem.
\begin{theorem}\label{thm_main}
For $T\ge 2$, we have
$\rho(\bm {Q}^{(T)}_{\mathrm{ch}}) < \rho(\bm Q^{(T)}_{\mathrm{const}})$. 
\end{theorem}
The proof can be found in App.~\ref{app_2} {in Supplemental Material}. 
This indicates that the convergence speed of 
CHGD is strictly faster than that of GD with the optimal step size when we compare every $T$ iteration. 


The convergence rate of GD is defined as 
$
R:=\liminf_{t\rightarrow\infty} \rho(\bm Q^{(t)})^{1/t}
$
 using the spectral radius per iteration. 
From (\ref{eq_qt}), the convergence rate $R_{\mathrm{ch}}(T)$ of GD with the Chebyshev steps (CHGD) of length $T$
 is bounded by
\begin{equation}
R_{\mathrm{ch}}(T) \le\left[\mathrm {sech}\left(T \cosh^{-1}\left(\frac{\kappa+1}{\kappa-1}\right) \right) \right]^{\frac{1}{T}}, \label{eq_ch_rho}
\end{equation}
which is lower than the convergence rate $R_{\mathrm{s}}$ (\ref{eq_gd2}) of GD with the optimal constant step size. 
The rate $R_{\mathrm{ch}}(T)$ approaches the lower bound  $R_{*}\! =\! {(\sqrt\kappa-1)}/{(\sqrt\kappa+1)}$ of first-order methods from above.
In Figure~\ref{fig_poi1}, we show the convergence rates of GD and CHGD as functions of the condition number $\kappa$.
{For CHGD, some upper bounds of convergence rates with different $T$ are plotted.}
We confirmed that CHGD with $T\ge 2$ has a smaller convergence rate than GD with the optimal constant step size and the rate of CHGD approaches the lower bound (\ref{eq_ch_rho}) quickly as $T$ increases. {This means that CHGD is asymptotically, i.e., 
in the large-$T$ limit, optimal in terms of convergence rate.}

 \begin{figure}[t]
   \centering
   \includegraphics[width=0.88\hsize]{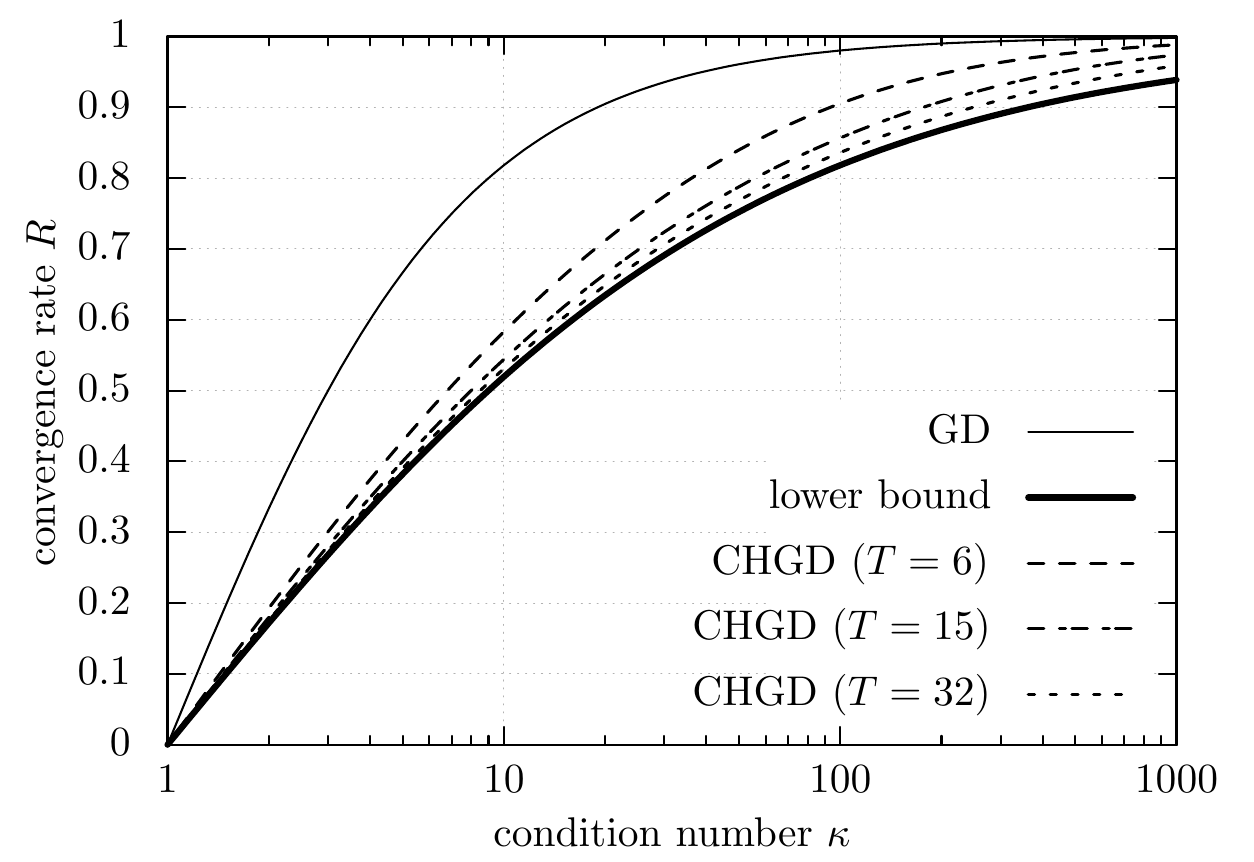}
    \caption{Convergence rates as functions of condition number $\kappa$. 
}
    \label{fig_poi1}
    \vskip -0.2in
\end{figure}

To summarize, in Sec.~\ref{sec_chst}-Sec.~\ref{sec_ca},
we introduced learning-free Chebyshev steps 
that minimize the upper bound of the spectral radius $\rho( \bm Q^{(T)})$ and improve the convergence rate.
{Additionally, we can show that the training process of DUGD also reduces the corresponding spectral radius (see App.~\ref{sec_ms}  in Supplemental Material).}
These imply that a learned step-size sequence of DUGD is possibly explained by the Chebyshev steps 
because Chebyshev steps are reasonable step sizes that reduce the spectral radius $\rho(\bm {Q}^{(T)})$.

\subsection{Numerical results}\label{sec_nc}

In this subsection, we examine DUGD following the setup in Sec~\ref{sec_dugdpre}. 
The main goal is to examine whether Chebyshev steps explain the nontrivial learned step-size parameters in DUGD.
Additionally, we also analyze the convergence property of DUGD and CHGD numerically and compare CHGD with other accelerated GDs.

\subsubsection{Experimental setting}
We describe the details of the numerical experiments.
DUGD was implemented using PyTorch 1.5~\cite{NIPS2019_9015}. 
Each training datum was given by a pair of random initial point $\bm {x}^{(0)}\in\mathbb{R}^n$ and optimal solution $\bm {x}^{\ast} = \bm{0}$.
A random initial point was generated as the i.i.d. Gaussian random vector with unit mean and unit variance.
The matrix $\bm {A}$ is generated by
$\bm {A}\!=\!\bm {H}^T\bm {H}$ {using $\bm {H}\in\mathbb{R}^{m\times n}$ whose elements independently follow $N(0,1/n)$}.
Then, 
{from the Marchenko-Pastur distribution law, as $n\rightarrow\infty$ with $m/n$ fixed to a constant, }
the maximum and minimum eigenvalues approach 
$(1+\sqrt{m/n})^2$ and $(1-\sqrt{m/n})^2$, respectively.
The matrix $\bm {A}$ was fixed throughout training process.
  
The training process was executed using incremental training.
See Sec.~\ref{sec_predugd} for details.
All initial values of $\{\gamma_t\}$ were set to $0.3$ unless otherwise noted. }
For each generation, the parameters were optimized to minimize the MSE loss function $L(\bm {x}^{(T)})$ between the output of DUGD and the optimal solution using $500$ mini batches of size $200$.
We used Adam optimizer~\cite{kingma2014adam} with a learning rate of $0.002$.

\subsubsection{Learned step sizes}

Next, we compared values of the learned step-size sequence with Chebyshev steps.
Figure~\ref{fig_step1} shows {examples of the sequences of length $T\!=\!6$ and $15$ when $(n,m)\!=\!(300,1200)$}. 
 {To compare parameters clearly, the learned step sizes and Chebyshev steps} were rearranged in descending order.
The lines with symbols represent the Chebyshev steps with 
$\lambda_n=9$ and $\lambda_1=1$. 
 Other symbols indicate the learned step-size parameters corresponding to five trials, that is, different matrices of $ \bm A$ and the training process with different random seeds.
 We found that the learned step sizes agreed with each other, thus indicating the self-averaging property of random matrices and success of the training process.
 More importantly, they were close to the Chebyshev steps, particularly when $\gamma$ was small.

\begin{figure}[t]
   \centering
   \includegraphics[width=0.88\hsize]{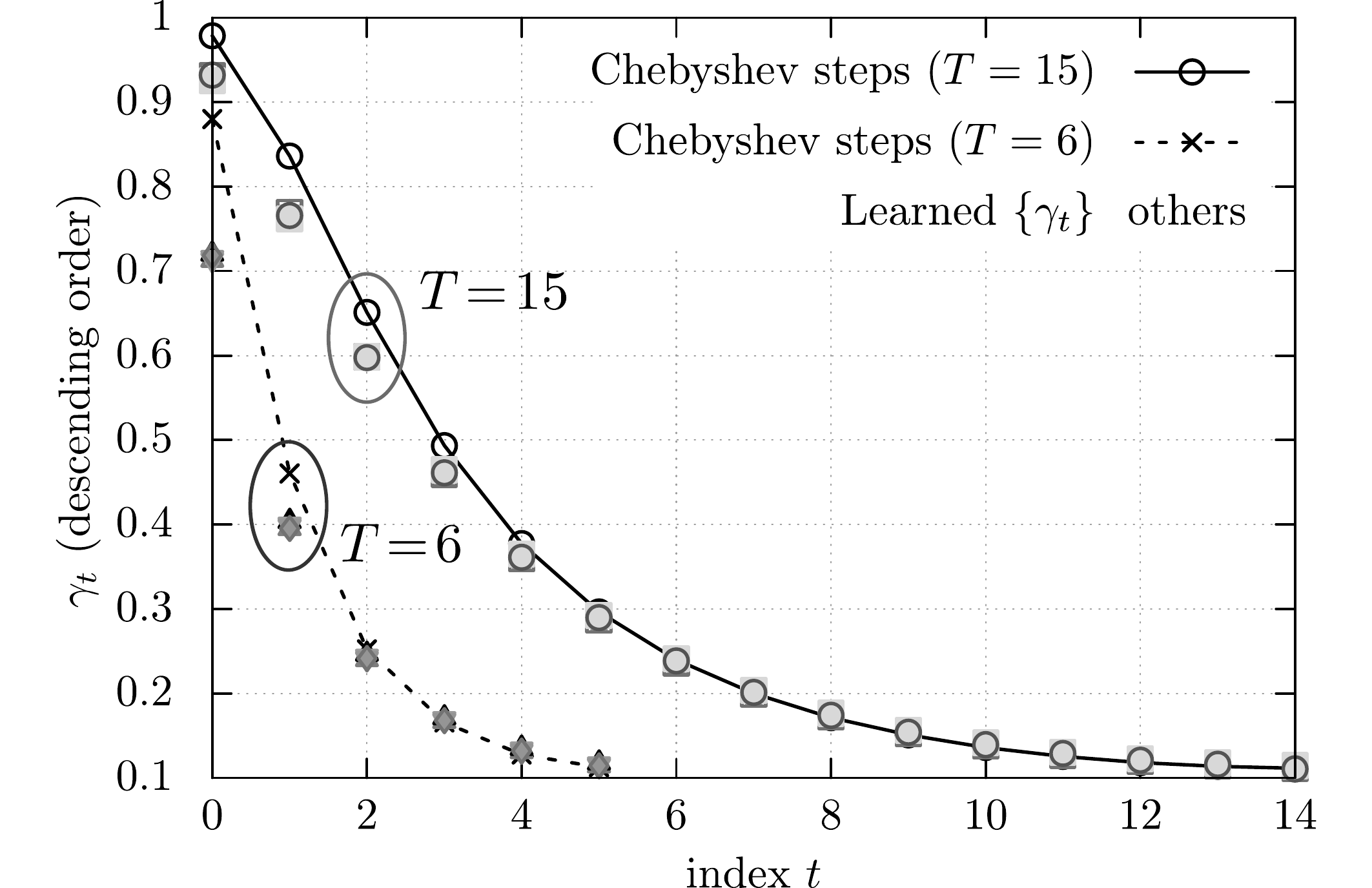}
    \caption{Chebyshev steps (open circles and cross marks) and learned step-size parameters in DUGD (others; $5$ trials) in descending order when $(n,m)=(300,1200)$ and $T=6$ (dotted) and $15$ (solid). 
}
    \label{fig_step1}
    \vskip -0.1in
\end{figure}

\subsubsection{Zig-zag shape}

 \begin{figure}[t]
 \vspace{-0.1in}
   \centering
   \includegraphics[width=0.88\hsize]{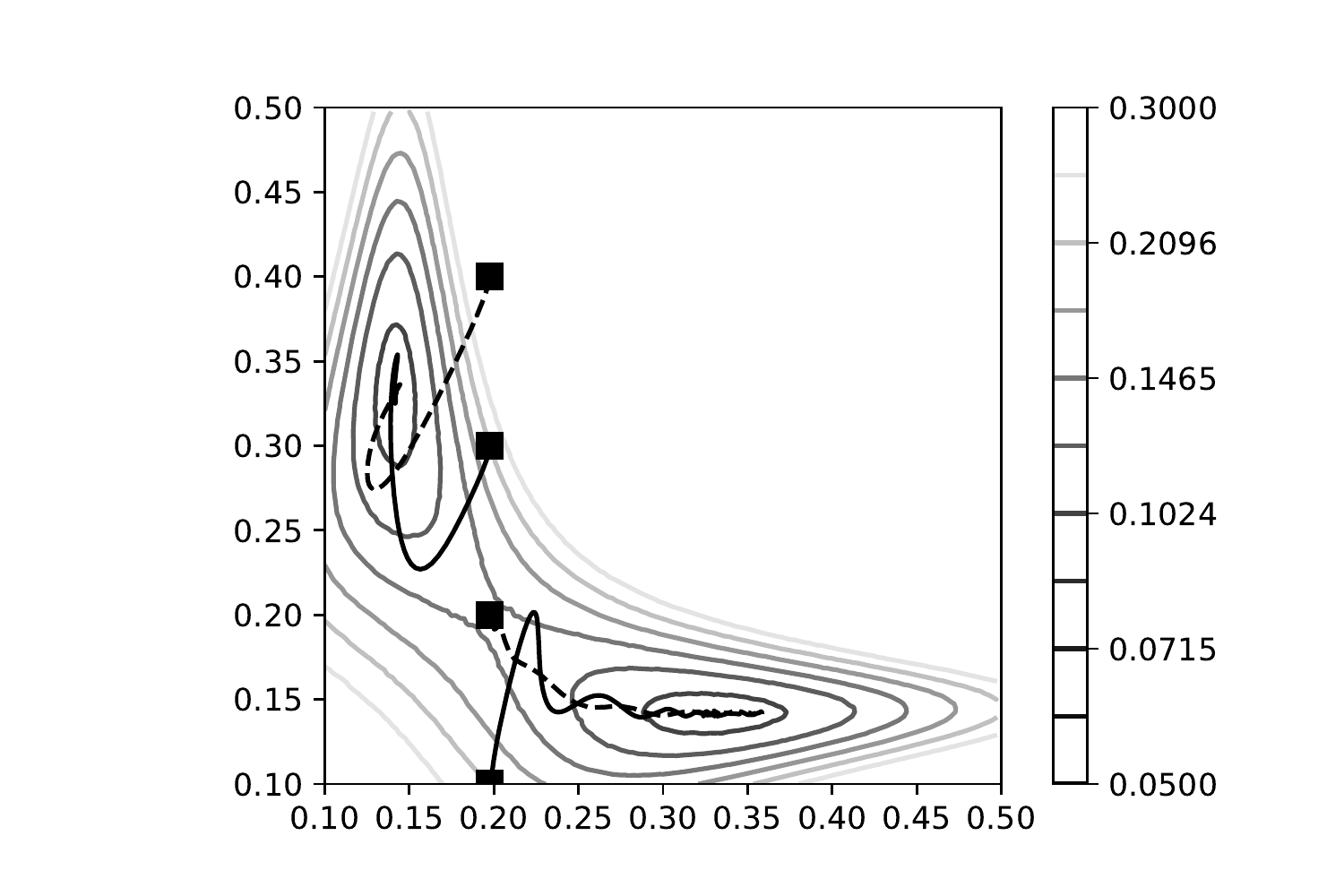}
    \caption{Trajectories $(\gamma_1,\gamma_2)$ in the second generation of incremental training from initial points $(\gamma_1^\ast,\gamma_2^{\mathrm{init}})$ (squares) where $\gamma_1^\ast$ is a trained value in the first generation.  
    Other lines show contours of the MSE loss 
    in the $\gamma_1$-$\gamma_2$ plane.
}
    \label{fig_dugt}
    \vskip -0.2in
\end{figure}

{The zig-zag shape of the learned step-size parameters is another nontrivial behavior of DUGD, although the order of parameters does not affect MSE performance after the $T$th iteration. 
It is numerically suggested that the shape depends on incremental training and initial values of $\{\gamma_t\}$.
Figure~\ref{fig_dugt} shows the trajectories of step sizes $(\gamma_1,\gamma_2)$ in the second generation of DUGD incremental training. 
Squares represent the initial point $(\gamma_1^\ast,\gamma_2^{\mathrm{init}})$ where $\gamma_1^\ast$ is a trained value of $\gamma_1$ in the first generation and $\gamma_2^{\mathrm{init}}=0.1,0.2,0.3,0.4$ is the initial value of $\gamma_2$.
We found that a convergent point clearly depends on the initial value of $\gamma_t$.  
This is natural because the optimal values of $(\gamma_1,\gamma_2)$ in the second generation show permutation symmetry.
Indeed, we confirmed that the location of a convergent point is based on the landscape of the MSE loss $\mathsf{E}_{\bm x^{(0)}}L(\bm x^{(2)}) $ used in the training process.

From this observation, we attempted to determine a permutation of the Chebyshev steps systematically by emulating incremental training.
Figure~\ref{fig_step2} shows the learned step-size parameters in DUGD ($T\!=\!11$) with different initial values of $\{\gamma_t\}$ and the corresponding permuted Chebyshev steps.
We found that they agreed with each other including the order of parameters.
Details of searching for such a permutation are provided in App.~\ref{app_c} {in Supplemental Material}.
}

\begin{figure}[t]
   \centering
   \includegraphics[width=0.88\hsize]{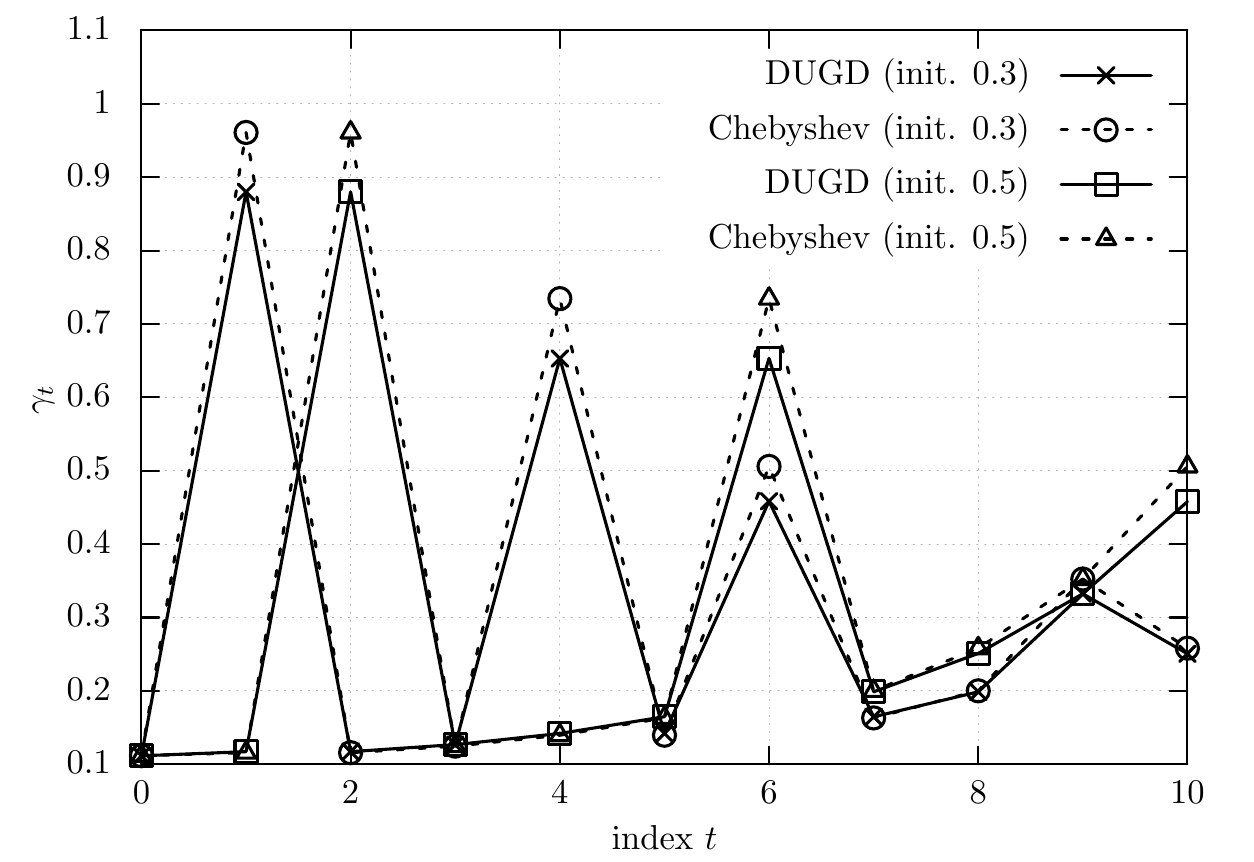}
    \caption{Zig-zag shape of the learned step-size parameters (solid lines) in DUGD and permuted Chebyshev steps (dotted lines) with different initial values of $\gamma_t$ when $(n,m)=(300,1200)$ and $T=11$. 
}
    \label{fig_step2}
    \vskip -0.1in
\end{figure}

\subsubsection{Difference between Chebyshev steps and learned step sizes}\label{sec_diff}

\begin{figure}[t]
   \centering
   \includegraphics[width=0.88\hsize]{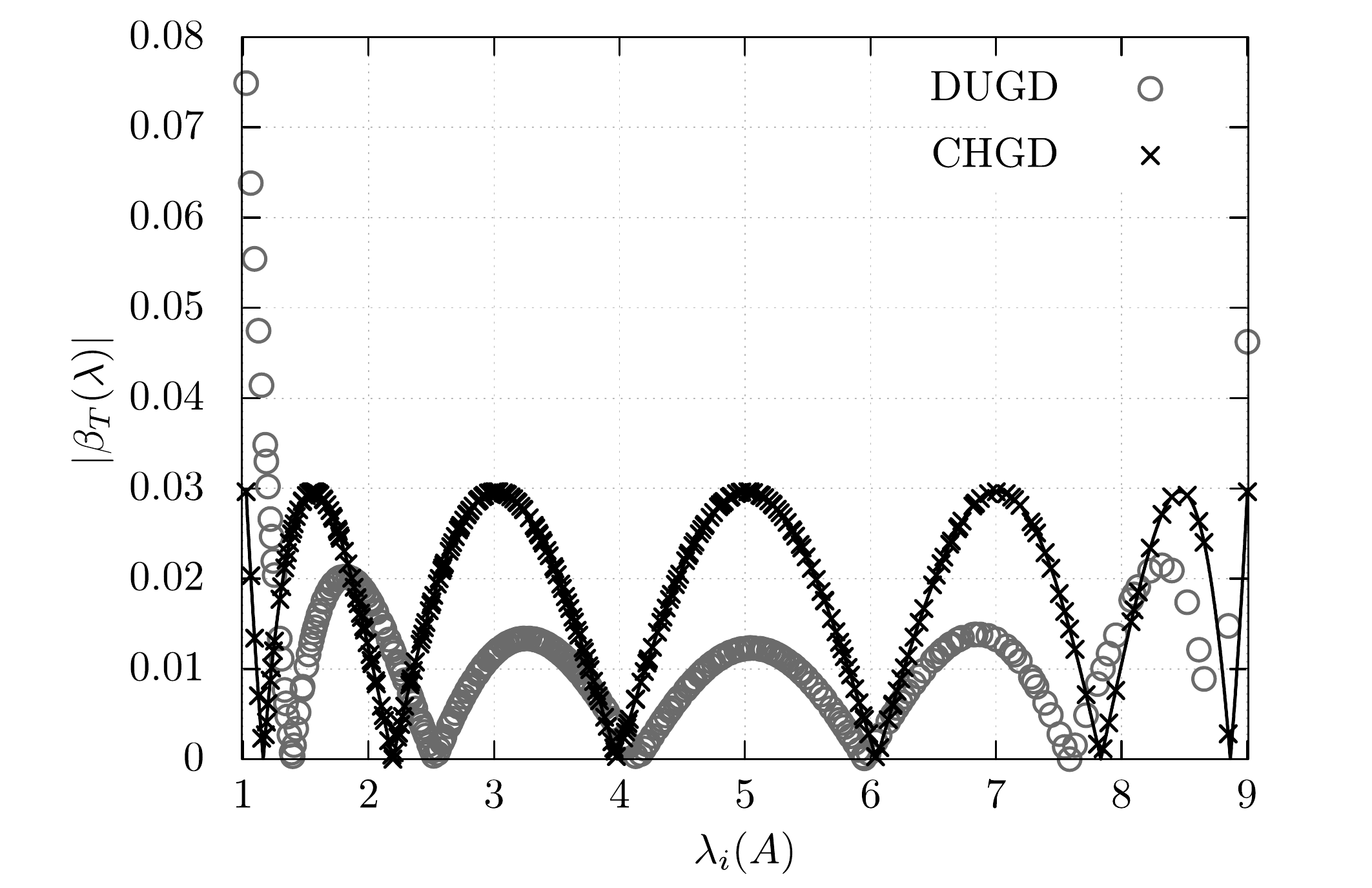}
    \caption{Absolute eigenvalues of $\bm {Q}^{(T)}$ using learned step-size parameters (circles) 
    as a function of eigenvalues $\lambda_i(\bm A)$ of $\bm {A}$ when $T=6$,  $\lambda_{min}(\bm A)=1.0$, and $\lambda_{max}(\bm A)=9.0$. 
    Cross marks represent the corresponding absolute eigenvalues when the Chebyshev steps are used.
}
    \label{fig_spec_c}
    \vskip -0.2in
\end{figure}

Figure~\ref{fig_spec_c}.
shows absolute eigenvalues $|\beta_T(\lambda_i(\bm A))|$ of ${Q}^{(T)}$ as a function of eigenvalue $\lambda_i(\bm A)$ in 
In the figure, circles represent $|\beta_T(\lambda_i(\bm A))|$ using learned step-size parameters in DUGD, whereas
cross marks are $|\beta_T^{ch}(\lambda_i(\bm A))|$ using Chebyshev steps when $T\!=\!6$, $\lambda_{min}(\bm A)\!=\!1.0$, and $\lambda_{max}(\bm A)\!=\!9.0$.
We found that $\{|\beta_T(\lambda_i(\bm A))|\}$ of DUGD were smaller than
 those of the Chebyshev steps in the high spectral-density regime and were larger otherwise.
This is because it reduced the MSE loss that all eigenvalues of the matrix $\bm A$ contributed.
By contrast, it increased the maximum value of $|\beta_T(\lambda_i(\bm A))|$ corresponding
 to the spectral radius $\rho(\bm  Q^{(T)})$. 
In this case, the spectral radius of DUGD was $0.074$, whereas that of the Chebyshev steps was $0.029$. 
Note that the spectral radius using the optimal constant step size $\gamma_{\mathrm s}=1/5$ was $\rho(\bm  Q_{\mathrm const}^{(T)})\simeq 0.262$.
We found that DUGD accelerated the convergence speed in terms of the spectral radius, whereas the
Chebyshev steps further improved the convergence rate.

\subsubsection{Performance analysis and convergence rate}

{Here, we compare the performance of DUGD and CHGD with the convergence rate evaluated in Section~\ref{sec_ca}.}

\begin{figure}[t]
   \centering
   \includegraphics[width=0.88\hsize]{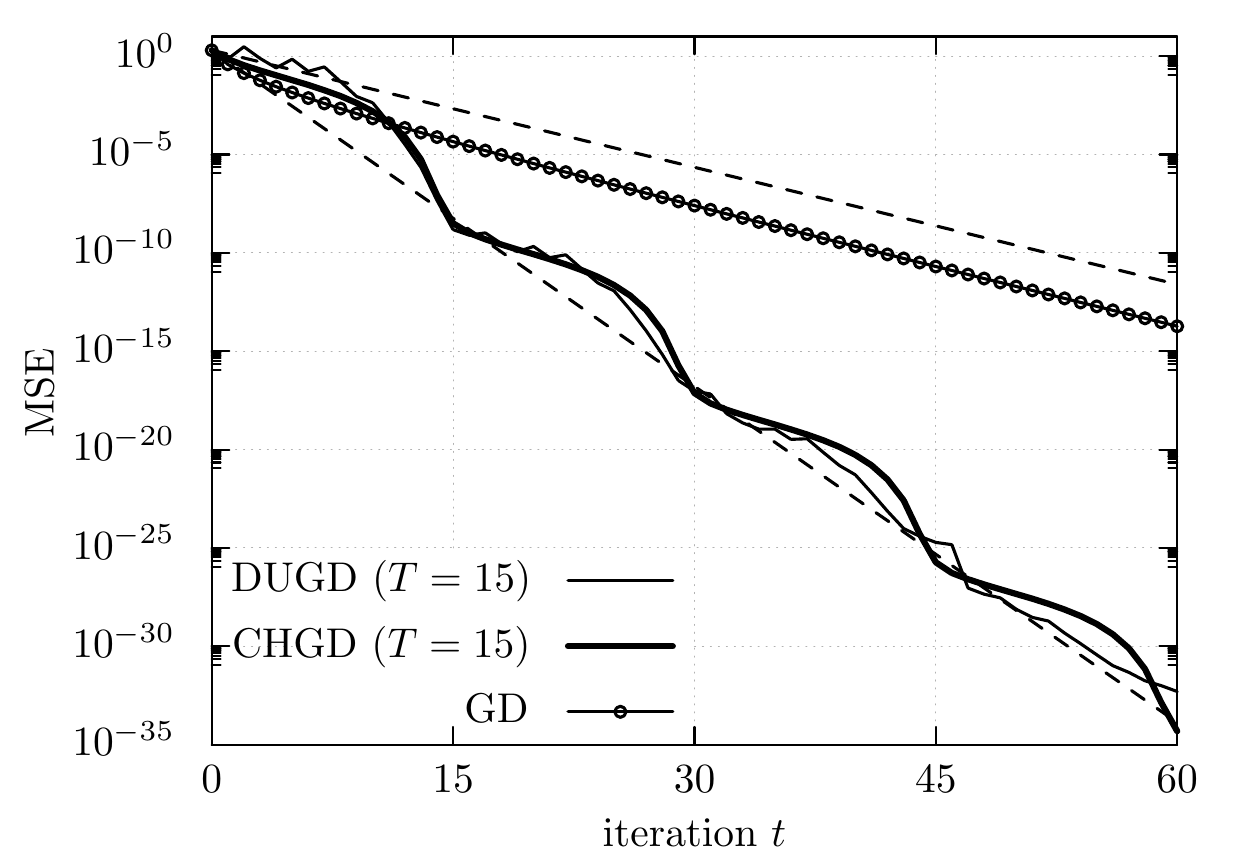}
    \caption{Comparison of the MSE of GD algorithms when $(n,m)=(300,1200)$ ($\kappa=8.79$).
    Dashed lines represent slopes of the convergence rate of CHGD (\ref{eq_ch_rho}) (lower) and GD (\ref{eq_gd2}) (upper).
}
    \label{fig_perf}
    \vskip -0.1in
\end{figure}

In the experiment 
DUGD  was trained the step sizes with $T=15$.
CHGD was executed repeatedly with the Chebyshev steps of length $15$.
{Figure~\ref{fig_perf} shows the MSE performance of DUGD, CHGD, and GD with the optimal constant step size when $(n,m)=(300,1200)$.
We found that DUGD and CHGD converged faster than GD, indicating that a proper step-size parameter sequence accelerated the convergence speed.
Although DUGD had slightly better MSE performance than CHGD when $t=15$, 
CHGD exhibited faster convergence as the number of iterations increased.
This is because the spectral radius of CHGD was smaller than that of DUGD, as discussed in the last subsection.
Figure~\ref{fig_perf} also shows the MSE calculated by (\ref{eq_dumse}) using the convergence rates.
We found that the upper bound of the convergence rate (\ref{eq_ch_rho}) correctly predicted the convergence properties of CHGD.}

\subsubsection{Comparison with accelerated GD}\label{sec_coGD}

\begin{figure}[t]
   \centering
   \includegraphics[width=0.88\hsize]{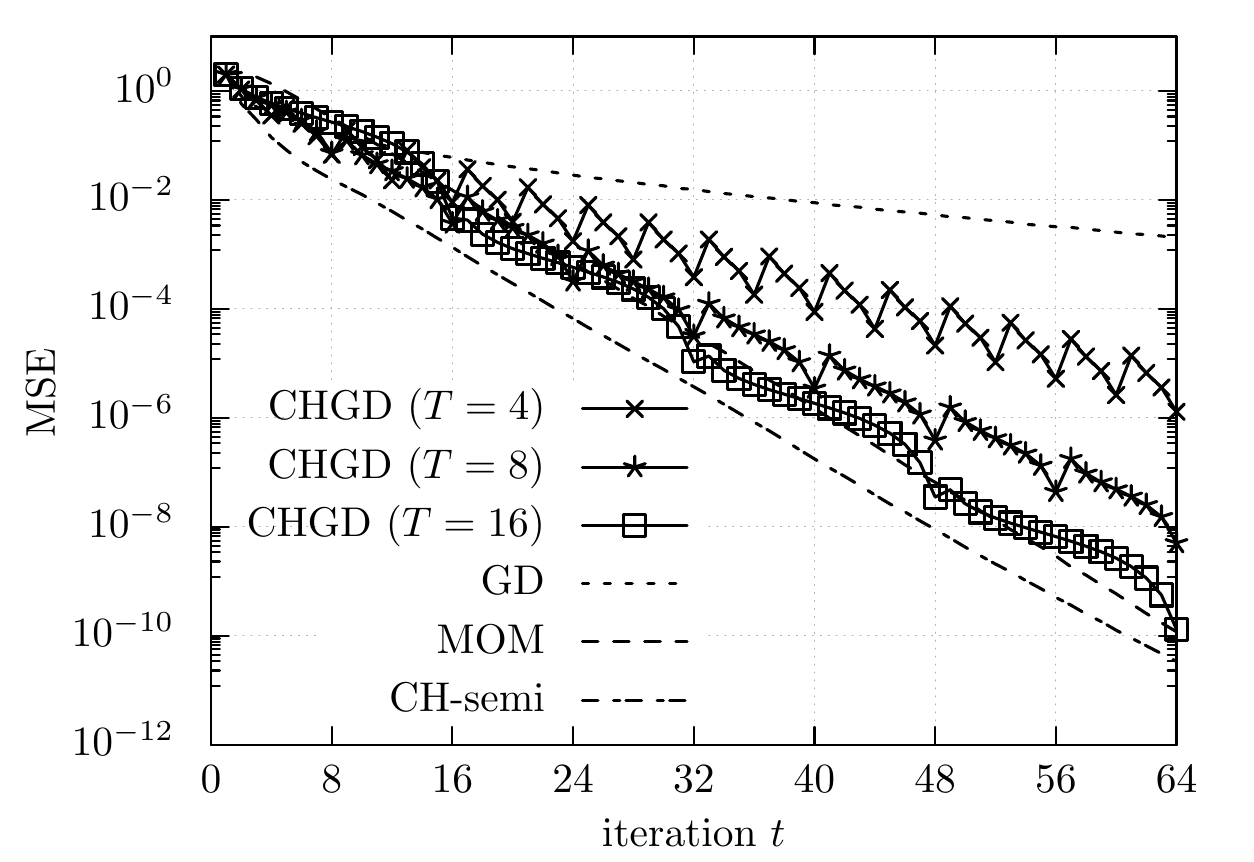}
    \caption{MSE performance of CHGD ($T\!=\!4,8,16$) and other GD algorithms when $(n,m)\!=\!(300,450)$ ($\kappa\!=\!88.1$, $10^4$ samples).
}
    \label{fig_comp}
    \vskip -0.2in
\end{figure}

To demonstrate the convergence speed of CHGD, 
we compared CHGD with GDs with a momentum term~\footnote{A sample code is available at \url{https://github.com/wadayama/Chebyshev}.}: 
the momentum (MOM) method (\ref{eq_mom}) and Chebyshev semi-iterative (CH-semi) method (\ref{eq_chsemi}).

Figure~\ref{fig_comp} shows the MSE performance of CHGD ($T\!=\!4,8,16$) and those of other GD algorithms when $(n,m)=(300,1200)$. 
We found that CHGD improved its MSE performance as $T$ increased. In particular, CHGD ($T=16$) showed reasonable performance compared with MOM and CH-semi. 
This indicates that CHGD exhibits acceleration of the convergence speed comparable to MOM and CH-semi
when $T$ is sufficiently large.
{We can conclude that CHGD is an \emph{accelerated GD algorithm without momentum terms}, which does not require a training process like DUGD.} 




\section{Extension of Theory of Chebyshev Steps}\label{sec_fpi}


In the previous section, we saw that Chebyshev steps successfully accelerate 
the convergence speed of GD. The discussion 
can be straightforwardly extended to local convergence analysis of 
the strongly convex objective functions 
because {such} a 
function can be well approximated by a quadratic function around the optimal point.

In this section, we present a novel method termed {\em Chebyshev-periodical successive over-relaxation (PSOR)}
for accelerating the convergence speed of
linear/nonlinear fixed-point iterations based on Chebyshev steps.

\subsection{Preliminaries}

In this subsection, we briefly summarize some background materials required for 
this section.
\subsubsection{Fixed-point iterations}
A {\em fixed-point iteration} 
\begin{equation}
\bm x^{(k+1)} = f(\bm x^{(k)}), \ k = 0, 1, 2, \ldots 
\end{equation}
 is widely used for 
solving scientific and engineering problems
such as inverse problems \cite{Bauschke11a}. 
 An eminent example is 
GD for minimizing a convex or non-convex objective function \cite{Boyed04}.
Fixed-point iterations are also widely employed for
solving large-scale linear equations.
For example, Gauss-Seidel methods, Jacobi methods, and 
conjugate gradient methods are
well-known fixed-point iterations~\cite{saad2003iterative}.
Another example of fixed-point iterations is the proximal gradient descent method 
{such as} ISTA~\cite{Daubechies04}
for sparse signal recovery problems. 
The projected gradient method also is an example that can be subsumed within the proximal gradient method.

\subsubsection{Successive over-relaxation}
Successive over-relaxation (SOR) is a well-known method for accelerating the Gauss-Seidel and Jacobi methods to solve
a linear equation $\bm P \bm x = \bm q$, where $\bm P \in \mathbb{R}^{n \times n}$ \cite{Young71}.
For example, Jacobi method is based on the following linear fixed-point iteration: 
\begin{equation}
	\bm x^{(k+1)} = f(\bm x^{(k)}) := \bm D^{-1} \left(\bm q - (\bm P - \bm D) \bm x^{(k)}  \right), \label{eq_jm}
\end{equation}
where $\bm D$ is a diagonal matrix whose diagonal elements are identical to the diagonal elements of $\bm P$.
{The update rule of} SOR for Jacobi method is simply given by
\begin{equation}
\bm x^{(k+1)} = \bm x^{(k)} + \omega_k\left( f( \bm x^{(k)} ) - \bm x^{(k)}  \right), 
\end{equation}
where $\omega_k \in (0, 2)$. 
A constant SOR factor $\omega_k := \omega_{const}$ is often employed in practice. 
When the SOR factors $\{\omega_k \}$ are appropriately chosen, the SOR-Jacobi method 
achieves much faster convergence compared with the standard Jacobi method.
SOR for linear equation solvers \cite{Young71} is particularly useful for 
solving sparse linear equations. Readers can refer to \cite{saad2003iterative} for a convergence analysis.

\subsubsection{Krasnosel'ski\v{\i}-Mann iteration}
Krasnosel'ski\v{\i}-Mann iteration \cite{mann1953mean,Bauschke11b} is a method similar to SOR.
For a non-expansive operator $\bm T$ in a Hilbert space, the fixed-point iteration of Krasnosel'ski\v{\i}-Mann iteration is given as
$
\bm x^{(k+1)} = \bm x^{(k)} + \lambda^{(k)} (\bm T \bm x^{(k)} - \bm x^{(k)}).
$
It is proved that the point $\bm x^{(k)}$ converges to a point in the set of fixed points of $\bm T$
if $\lambda^{(k)} \in [0,1]$ satisfies 
$
\sum_k \lambda^{(k)} (1 - \lambda^{(k)}) = +\infty.
$

\subsubsection{Proximal methods}

Proximal methods are widely employed for 
proximal gradient methods and projected gradient 
methods \cite{Parikh14} \cite{Combettes11}.
The proximal operator $\text{prox}_{h}(\cdot): \mathbb{R}^n \rightarrow \mathbb{R}^n $ 
for the function $h: \mathbb{R}^n \rightarrow \mathbb{R}^n$ is defined by
\begin{equation}
	\text{prox}_{h}(\bm v) := \arg \min_{\bm x\in\mathbb{R}^n} \left( h(\bm x) +  \frac 1 2 \|\bm x - \bm v\|^2 \right).
\end{equation}
Assume that we want to solve the following minimization problem:
\begin{equation} \label{objective}
	\text{minimize} \ u(\bm x) + v(\bm x),
\end{equation}
where $u: \mathbb{R}^n \rightarrow \mathbb{R}$ and $v: \mathbb{R}^n \rightarrow \mathbb{R}$.
The function $u$ is differentiable, and $v$ is a strictly convex function which is not necessarily differentiable.
The iteration of the proximal gradient method is given by
\begin{equation}
	\bm x^{(k+1)} = \text{prox}_{\lambda v}\left( \bm x^{(k)} - \lambda \nabla u(\bm x^{(k)}) \right),
\end{equation}
where $\lambda$ is a step size. It is known that $\bm x^{(k)}$ converges to the minimizer of (\ref{objective})
if $\lambda$ is included in the semi-closed interval $(0, 1/L]$, where $L$ is the Lipschitz constant of $\nabla u$.

\subsection{Fixed-point iteration}

Let $f: \mathbb{R}^n \rightarrow \mathbb{R}^n$ be a differentiable function.
Throughout the paper, we will assume that $f$ has a fixed point $\bm x^\ast\in\mathbb{R}^n$ satisfying 
$\bm x^\ast = f(\bm x^\ast)$ and that $f$ is {\em locally contracting mapping} around $\bm x^\ast$.
Namely, there exists $\epsilon ( > 0)$ such that
\begin{equation}
	\|f(\bm x) - f(\bm y) \|_2 < \|\bm x - \bm y\|_2
\end{equation}
holds for any $\bm x, \bm y \in B(\bm x^\ast, \epsilon)$ $(\bm x \ne\bm  y)$, where $B(\bm x^\ast, \epsilon)$ is  
$\epsilon$ ball defined by
$
B(\bm x^\ast, \epsilon) := \{\bm x \in \mathbb{R}^n: \|\bm x -\bm  x^\ast\|_2 \le \epsilon \}.	
$

These assumptions clearly show that the fixed-point iteration
\begin{equation} \label{iteration}
\bm x^{(k+1)} = f(\bm x^{(k)}),\ k = 0, 1, 2, \ldots
\end{equation}
eventually converges to the fixed point $\bm x^\ast$ if the initial point 
$\bm x_0$ is included in $B(\bm x^\ast, \epsilon)$.

SOR is a well-known technique for accelerating the 
convergence speed for linear fixed-point iterations such as the Gauss-Seidel method \cite{Young71}.
The SOR (or Krasnosel'ski\v{\i}-Mann) iteration was originally defined for a linear update function $f$; 
however, it can be generalized for a nonlinear update function.
The SOR iteration is given by
\begin{eqnarray} \nonumber
	\bm x^{(k+1)} &=&\bm  x^{(k)} + \omega_k \left[f(\bm x^{(k)}) 
	- \bm x^{(k)} \right] \\ \label{SORiteration}
	&=& (1 - \omega_k)\bm x^{(k)} + \omega_k f(\bm x^{(k)}),
\end{eqnarray}
where $\omega_k (k = 0, 1, \ldots)$ is a real coefficient, which is
called the {\em SOR factor}. 
When $f$ is linear, and $\omega_k  = \omega_{const}$ is a constant, 
the optimal choice of $\omega$ in terms of the convergence rate is well-known \cite{saad2003iterative}; however,
the optimization of iteration-dependent SOR factors $\{\omega_k\}$ for nonlinear update functions
has not hitherto been studied to the best of the authors' knowledge.

We define the {\em SOR update function} $S^{(k)}: \mathbb{R}^n \rightarrow \mathbb{R}^n$ by
\begin{equation}
S^{(k)}(\bm x) := (1 - \omega_k) \bm x + \omega_k f(\bm x).
\end{equation}
Note that {$S^{(k)}(\bm x^\ast)=\bm x^\ast$}
holds for any $k$.
Because $\bm x^\ast$ is also the fixed point of $S^{(k)}$, 
the SOR iteration does not change the fixed point of the original fixed-point iteration.

\subsection{Periodical SOR}

Because $f$ is a differentiable function, 
it would be natural to consider the {\em linear approximation} around the fixed point $\bm x^\ast$. 
The Jacobian matrix of $S^{(k)}(\bm x)$ around $\bm x^\ast$
 is given by $\bm I_n - \omega_k (\bm I_n-\bm J^\ast)$.
Matrix $\bm J^\ast$ is the Jacobian matrix of $f$ at the fixed point $\bm x^\ast = (x_1^*,\ldots, x_n^*)^T$:
\begin{equation}
	\bm J^\ast := 
	\left(
	\begin{array}{ccc}
		\frac{\partial f_1}{\partial x_1}(x_1^\ast) &\frac{\partial f_1}{\partial x_2}(x_2^\ast) & \hdots \\	
		 \vdots & \ddots & \vdots \\	
		\frac{\partial f_n}{\partial x_1}(x_1^\ast) &\hdots & \frac{\partial f_n}{\partial x_n}(x_n^\ast) \\	
	\end{array}
	\right),
\end{equation}
where $\bm f := (f_1, f_2, \ldots, f_n)^T$.
In the following discussion, we use the abbreviation for simplicity:
\begin{equation}
	\bm B :=\bm  I_n - \bm J^\ast.
\end{equation}
{Here, we assume that $\bm B$ is a Hermitian matrix.}

The function $S^{(k)}(\bm x)$ in the SOR iteration (\ref{SORiteration})
can be approximated using
\begin{equation}
	S^{(k)}(\bm x) \simeq S^{(k)}(\bm x^\ast) +  \left(\bm I_n - \omega_{k}\bm  B \right) (\bm x - \bm x^\ast) 
\end{equation}
because of Taylor series expansion around the fixed point.
In this context, for expository purposes, we will omit the residual terms such as $o(\|\bm x - \bm x^\ast\|_2)$ appearing in the
Taylor series expansion. A detailed argument which considers the residual terms can be found in the proof of Thm~\ref{convergence_rate}.

Let $\{\bm x^{(0)}, \bm x^{(1)}, \ldots \}$ be a point sequence obtained by a SOR iteration.
By letting $\bm x = \bm x^{(k)}$, we have
\begin{equation} \label{aprox}
	\bm x^{(k+1)} - \bm x^\ast \simeq  \left(\bm I_n - \omega_{k} \bm B \right) (\bm x^{(k)} - \bm x^\ast).
\end{equation}

In the following argument in this paper, we assume 
{\em periodical SOR (PSOR) factors} $\{\omega_k \}_{k=0}^{T-1}$ satisfying 
\begin{equation}
\omega_{\ell T + j} = \omega_j \quad (\ell = 0,1,2,\ldots,\  j = 0, 1, 2, \ldots, T-1),
\end{equation}
where $T$ is a positive integer.	
SOR utilizing PSOR factors are called a {\em PSOR}.

Applying a norm inequality to (\ref{aprox}), we can obtain
\begin{equation} \label{eq_fpi_in}
	\|\bm x^{((\ell + 1)T)} - \bm x^\ast \|_2  \le \rho\left(\prod_{k = 0}^{T-1} \left(\bm I_n - \omega_{k} \bm B \right) \right)  \|\bm x^{(\ell T)} - \bm x^\ast\|_2
\end{equation}
for sufficiently large $\ell$. 
The inequality (\ref{eq_fpi_in}) shows that the local convergence behavior around the fixed point 
is dominated by
the spectral radius of {$\bm W^{(T)}:= \prod_{k = 0}^{T-1}(\bm I_n - \omega_{k} \bm B)$}.

If the spectral radius satisfies  {\em spectral condition}  {$\rho_T := \rho(\bm  W^{(T)} ) < 1$,} 
we can expect local linear convergence around the fixed point.

\subsection{Chebyshev-PSOR factors}

As assumed in the previous subsection, $\bm B$ is a Hermitian matrix, and 
we can show that $\bm B$ is also a positive definite matrix due to the assumption that $f$ is contractive mapping.
Therefore, we can apply the theory of Chebyshev steps in the last section by replacing the matrix $\bm A$ in Sec.~\ref{sec_chst} to $\bm B$.  
We define the {\em Chebyshev-PSOR factors} $\{\omega_k^{ch}\}_{k=0}^{T-1}$ by
\begin{equation} \label{SORfactors}
\omega_k^{ch}
    :=\left[ \lambda_+(\bm B) + \lambda_-(\bm B) \cos \left(\frac{2k+1}{2T} \pi \right) \right]^{-1},
\end{equation}
where $\lambda_\pm (\cdot)$ are defined by (\ref{eq_lam1}). 
We will call the PSOR employing $\{\omega_k^{ch}\}_{k=0}^{T-1}$ the {\em Chebyshev-PSOR}.

The asymptotic convergence rate of the Chebyshev-PSOR can be evaluated as in Sec.~\ref{sec_ca}.
The following theorem clarifies the asymptotic local convergence behavior of a Chebyshev-PSOR.
\begin{theorem} \label{convergence_rate}
Assume that $\bm B$ is a positive definite Hermitian matrix.
Suppose that there exists a convergent point sequence $\{\bm x^{(0)}, \bm x^{(1)},\ldots \}$ produced  
by a Chebyshev-PSOR, which is convergent to the fixed point $\bm x^*$.
The following inequality 
\begin{equation} \label{rate_ineq}
	\lim_{\ell \rightarrow \infty}\frac{\|\bm x^{((\ell+1)T)} - \bm x^*\|_2}{\|\bm x^{(\ell T)} - \bm x^*\|_2}	
	\le  \mathrm{sech}\left(T \cosh^{-1}\left(\frac{\kappa+1}{\kappa-1}\right) \right) 
\end{equation}
holds, where $\kappa := \kappa(\bm B)$. 
\end{theorem}

{See App.~\ref{app_thm3} in Supplemental Material for the proof.}
The right-hand side of (\ref{rate_ineq}) 
{is smaller than one} because $\mathrm{sech}(x)< 1$ holds for $x>0$.
Thus, we can expect local linear convergence around the fixed point.

\subsection{Convergence acceleration by Chebyshev-PSOR}

In the previous subsection, we observed that the spectral radius $\rho_T$ is 
strictly smaller than one for any Chebyshev-PSOR.
We still need to confirm whether the Chebyshev-PSOR 
provides definitely faster convergence compared with the original fixed-point iteration.

Around the fixed point $\bm x^*$, the original fixed-point iteration (\ref{iteration})
achieves 
$\|\bm x^{(k)} - \bm x^*\|_2\le q_{org}^k \|\bm x^{(0)} -\bm  x^*\|_2$, where $q_{org} := \rho(\bm J^*) < 1$.
For comparative purposes, similar to (\ref{eq_ch_rho}), we define the convergence rate of the Chebyshev-PSOR as
\begin{equation}
q_{ch}(T):=
\left[ \mathrm{sech}\left(T \cosh^{-1}\left(\frac{b+a}{b-a}\right) \right) \right]^{\frac 1 T},
\end{equation}
where $a=\lambda_{min}(\bm B),b=\lambda_{max}(\bm B)$. The second line is derived from (\ref{eq_ch1}) and (\ref{eq_qt}).
From Thm.~\ref{convergence_rate}, 
\begin{equation}
\|\bm x^{(k)} - \bm x^\ast\|_2 < q_{ch}^k \|\bm x^{(0)} - \bm x^\ast\|_2	
\end{equation}
holds if $k$ is a multiple of $T$ and $\bm x^{(0)}$ is sufficiently close to $\bm x^\ast$.
We show that $q_{ch}(T)$ is a bounded monotonically decreasing function.
The limit value of $q_{ch}(T)$ is given by 
\begin{equation}
q_{ch}^\ast := \lim_{T \rightarrow \infty} q_{ch}(T) = 
\exp\left(-\cosh^{-1}\left(\frac{b+a}{b-a}\right)  \right),
\end{equation}
using 
$\lim_{x \rightarrow \infty}\left[ \mathrm{sech}(\alpha x) \right]^{1/x} = e^{-\alpha}.$
In the following discussion, we assume 
that $\lambda_{min}(\bm J^\ast) = 0$ and $\lambda_{max}(\bm J^\ast) = \rho_{org} (0 < \rho_{org} < 1)$.
In this case, we have $a := 1- \rho_{org}$ and $b := 1$.
We thus have, for any $T$,
\begin{equation}
q_{ch}^\ast < q_{ch}(T) < \rho_{org}.
\end{equation}
Figure \ref{xi_function} presents the region where $q_{ch}(T)$ can exist.
The figure implies that the Chebyshev-PSOR can definitely accelerate the 
local convergence speed if 
$\bm x_0$ is sufficiently close to $\bm x^*$.

\begin{figure}[t]
\begin{center}
\centerline{\includegraphics[width=0.88\hsize]{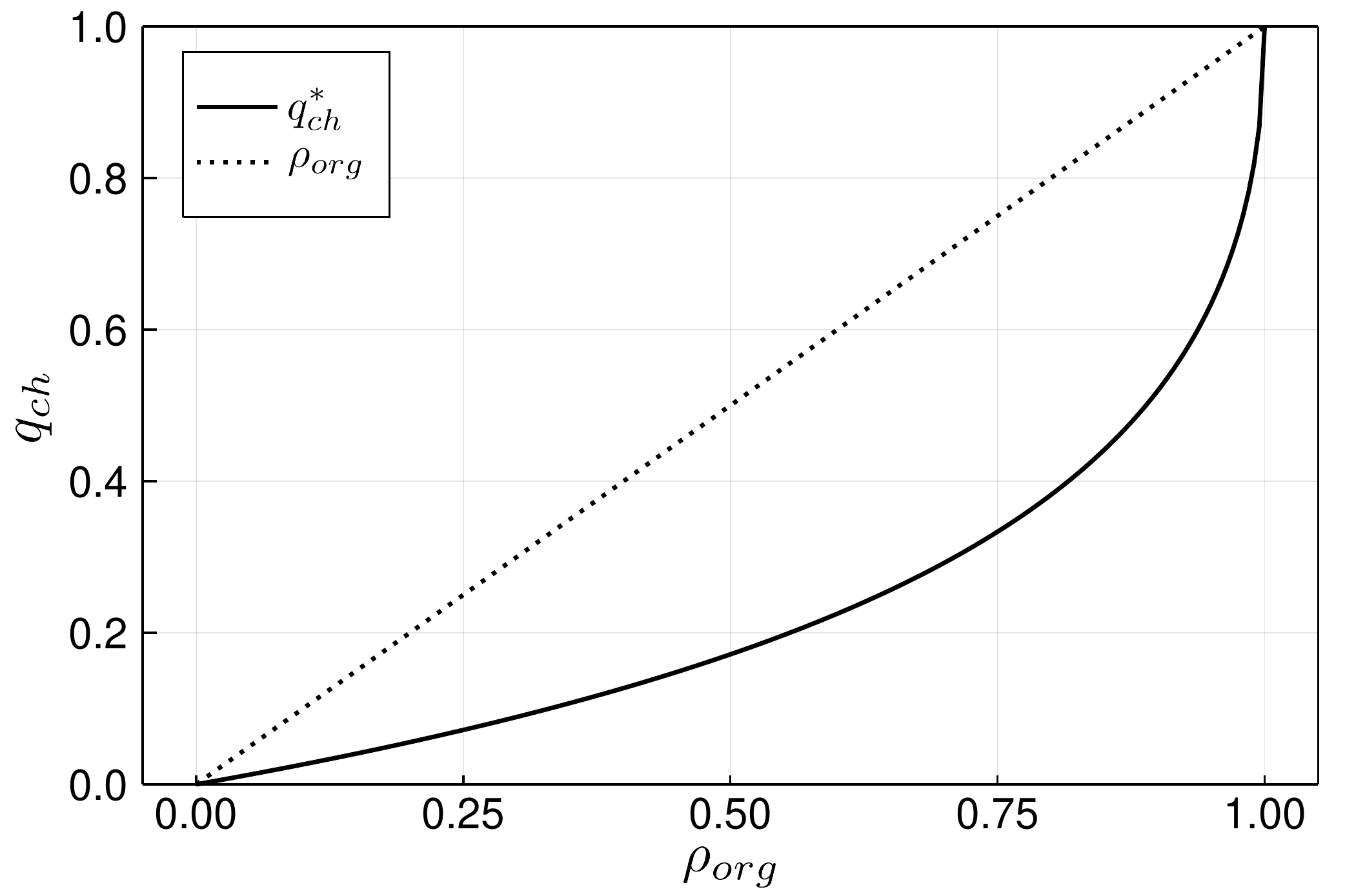}}
\caption{The convergence rate region of $q_{ch}(T)$: the straight line represents $\rho_{org}$ 
and the curve corresponds to $q_{ch}^*$.
If $\bm x_0$ is sufficiently close to $\bm x^*$, $q_{ch}(T)$ falls in the area surrounded by 
the line and the curve.}
\label{xi_function}
\end{center}
\vskip -0.2in
\end{figure}

\subsection{Affine composite update function}

In this subsection, we discuss an important special case where 
the update function $f$ has the form 
$
f(\bm x) := g(\bm A \bm x + \bm b)		
$
which is a composition of the affine transformation $\bm A \bm x + \bm b$ and 
a component-wise nonlinear function $g$ where 
$\bm A \in \mathbb{R}^{n \times n}, \bm b \in \mathbb{R}^n$.
Note that a function $g:\mathbb{R} \rightarrow \mathbb{R}$ is applied to 
$\bm x \in \mathbb{R}^n$ in a component-wise manner as
$
	g(\bm x) := (g(x_1), \ldots, g(x_n))^T.	
$

The fixed-point iteration based on such an affine composite update function 
is given by 
\begin{equation} \label{affine}
\bm x^{(k+1)} = g( \bm A \bm x^{(k)} + \bm b), \ k = 0, 1,2,  \ldots.
\end{equation}
Such a fixed-point iteration is practically important 
because 
many iterative optimization methods such as proximal gradient and 
projected gradient methods
are included in this class.

\subsubsection{Eigenvalues of Jacobian matrix}

The Jacobian matrix of the update function $f(\bm x) = g( \bm A \bm x+ \bm b)$ at the fixed point $x^*$ 
is given by 
$
	\bm J^* = \bm Q^* \bm A,		
$
where $\bm Q^* := \{q^*_{i,j}\} \in \mathbb{R}^{n \times n}$ is a diagonal matrix with
the diagonal elements defined by
\begin{equation}
	q^*_{i,i} = g'\left(\sum_{j=1}^n a_{i,j} x_j^* + b_i \right),\ i = 1,2,\ldots,n.
\end{equation}
Note that $\bm Q^* \bm A$ is not necessarily symmetric
even if $\bm A$ is symmetric.

Even though the matrix $\bm Q^* \bm A$ is not symmetric in general, all the eigenvalues of $\bm Q^* \bm A$
are real if certain conditions are met.
The following theorem provides a sufficient condition.
\begin{theorem} \label{real_th}
	Assume that $g: \mathbb{R} \rightarrow \mathbb{R}$ is differentiable,
	 and $g'(x) \ge 0$ holds for any $x \in \mathbb R$.
	If $\bm A \in \mathbb{R}^{n \times n}$ is a symmetric non-singular matrix,  
	all eigenvalues of $\bm J^* = \bm Q^* \bm A$ are real.
\end{theorem}
{The proof is based on~\cite{Drazin62} and found in the App.~\ref{app_thm4} in Supplemental Material.}

\subsubsection{Non-symmetric $\bm B$}

As shown, all eigenvalues of $\bm J^*$ are real, but $\bm J^*$ is not symmetric in general.
Recall that Thm.~\ref{convergence_rate} can be applied to the cases where $\bm J^*$ is symmetric, i.e., $\bm J^*$ is Hermitian.
This means that we cannot directly use Thm.~\ref{convergence_rate} for convergence analysis.
The Hermitian assumption is important to obtain the last equality in (\ref{long_ineq}) {in Supplemental Material}.

We now consider a case where $\bm B$ is not symmetric, but all the eigenvalues of $\bm B$ are real.
{Although that the proof of Thm.~\ref{convergence_rate} is based on $\rho(\bm W^{(T)}) = \|\bm W^{(T)}\|_2$, the relation does not hold for this case. Instead, an inequality $\rho(\bm W^{(T)}) \le \| \bm W^{(T)}\|_2$ holds.}
For non-symmetric (i.e., non-Hermitian) $\bm B$, the Chebyshev-PSOR can be regarded as a method 
to reduce the spectral radius $\rho_T$ which is a lower bound of $\| \bm W^{(T)}\|_2$.

\subsection{Numerical experiments}

\subsubsection{Linear fixed-point iteration: Jacobi method}
\begin{figure}[t!]
\begin{center}
\centerline{\includegraphics[width=0.88\hsize]{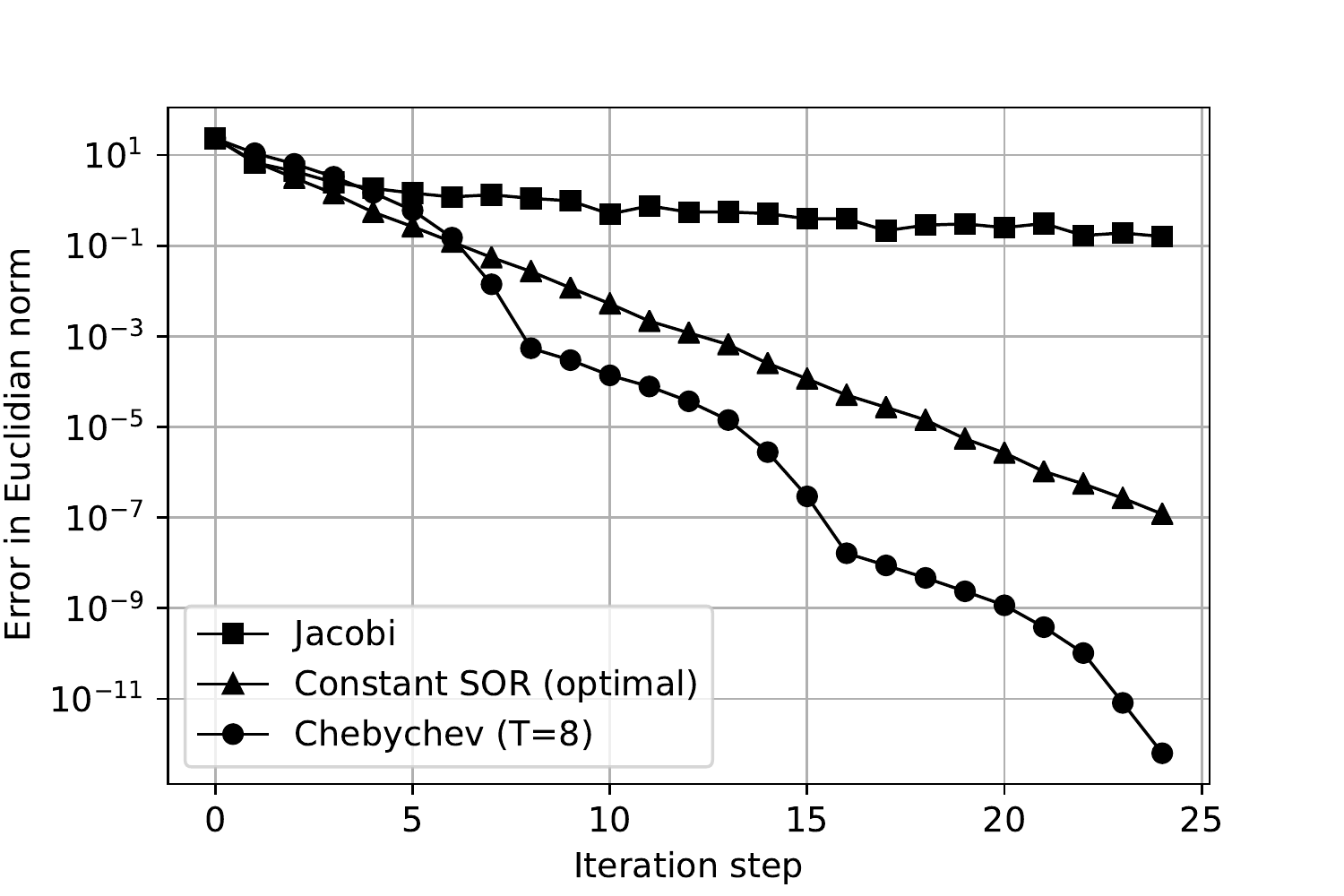}} 
\caption{Jacobi method, constant SOR factor, and Chebyshev-PSOR: 
The linear equation $\bm P \bm x = \bm 0$ is assumed without loss of generality where
$\bm P = \bm I_n + \bm M^T \bm M \in \mathbb{R}^{512 \times 512}$ is a diagonally dominant matrix.
Each element of $\bm M$ follows ${\cal N}(0, 0.03^2)$. The minimum and maximum 
eigenvalues of $\bm B = \bm I_n - \bm A$ are given by 
$\lambda_{max}(\bm B) = 1.922$ and $\lambda_{min}(\bm B) = 0.6766$, respectively.
The three curves represent errors in the Euclidean norm from the fixed point, i.e., $\|\bm x^{(k)} - \bm x^*\|_2$ where
$\bm x^* = \bm 0$.
The optimal constant SOR factor 
is given by $\omega_{SOR} = 2/(\lambda_{min}(\bm B) + \lambda_{max}(\bm B)) = 0.7697$ \cite{saad2003iterative}.
}
\label{jacobi}
\end{center}
\vskip -0.2in
\end{figure}

We can apply the Chebyshev-PSOR to Jacobi method { (\ref{eq_jm})} for accelerating the convergence~\footnote{A sample code is available at \url{https://github.com/wadayama/Chebyshev}.}.
Figure \ref{jacobi} shows the convergence behaviors.
It should be noted that the optimal constant SOR factor is 
exactly the same as the Chebyshev-PSOR factor with $T=1$.
We can see that the Chebyshev-PSOR provides much faster convergence 
compared with the standard Jacobi and the SOR with the constant optimal SOR factor.
The error curve of the Chebyshev-PSOR follows a wave-like shape.
This is because the error is tightly bounded periodically as shown in Thm.~\ref{convergence_rate}; i.e., 
the error is guaranteed to be small when the iteration index is a multiple of $T$.

\subsubsection{Nonlinear fixed-point iterations}

A nonlinear function $f = (f_1, f_2)^T: \mathbb{R}^2 \rightarrow \mathbb{R}^2$,
\begin{eqnarray}
f_1(x_1, x_2) &:=& x_1^{0.2} + x_2^{0.5}, \\
f_2(x_1, x_2) &:=& x_1^{0.5} + x_2^{0.2}
\end{eqnarray}
is assumed here.
Figure \ref{polynomial} (left) shows errors as a function of
the number of iterations. We observe that the Chebyshev-PSOR accelerates the convergence to the
fixed point. The Chebyshev-PSOR results in a zig-zag shape of the PSOR factors as 
depicted in Fig. \ref{polynomial} (right).
\begin{figure}[t!]
\begin{center}
\centerline{\includegraphics[width=0.88\hsize]{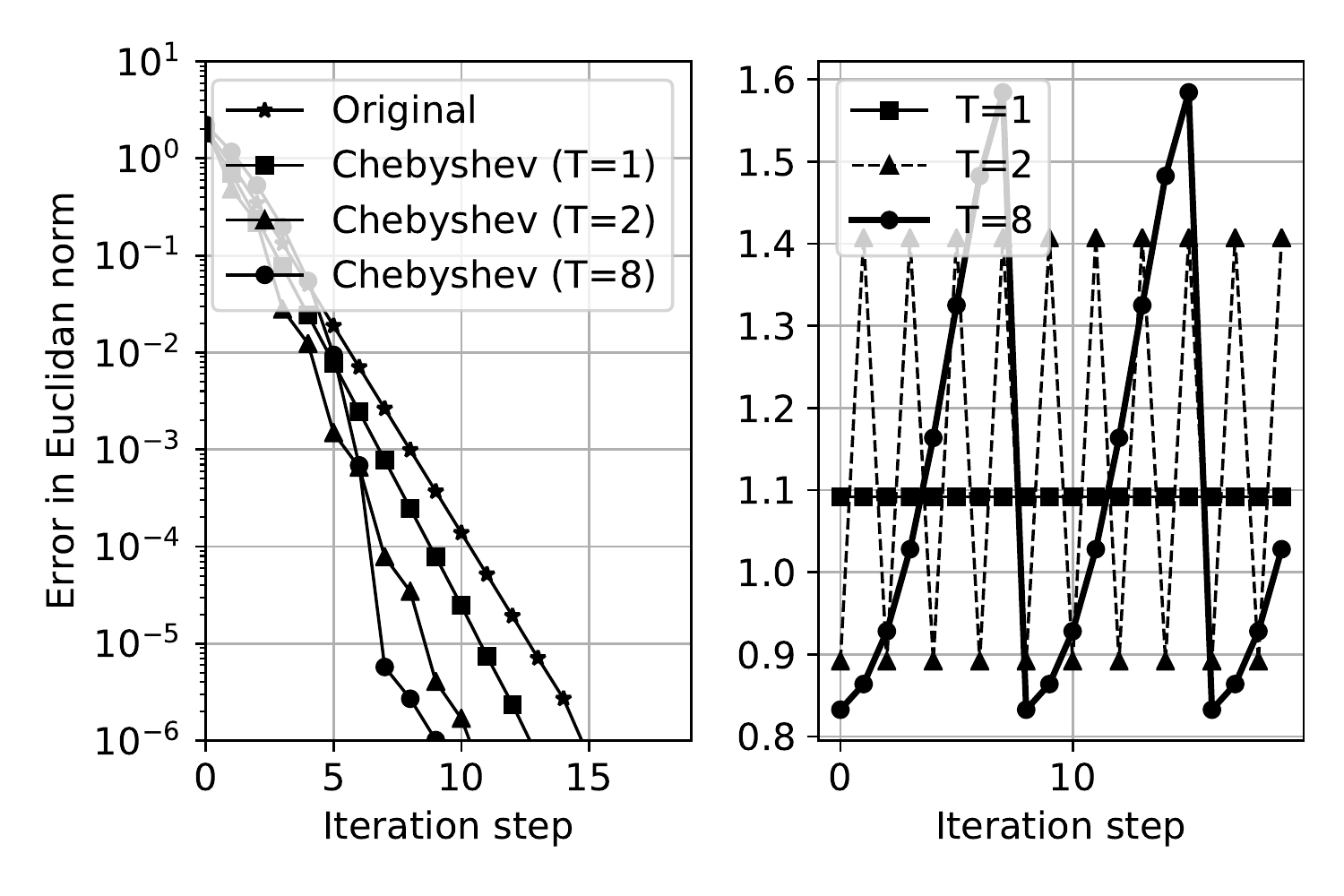}}
\caption{Nonlinear fixed-point iteration: 
The left figure shows the error $\|\bm x^{(k)} - \bm x^*\|_2$ for 
the original fixed-point iteration and the Chebyshev-PSORs $(T= 1,2,8)$.
The fixed point is $\bm x^* = (2.96, 2.96)^T$.
The minimum and maximum eigenvalues of $\bm B = \bm I_n - \bm J^*$ are 
$\lambda_{max}(\bm B) = 1.216$ and $\lambda_{min}(\bm B) = 0.626$, respectively.
The horizontal axes represent the iteration step. The right figure shows
the Chebyshev-PSOR factors for $T = 1, 2, 8$ as a function of the iteration step.
}
\label{polynomial}
\end{center}
\vskip -0.2in
\end{figure}

\begin{figure}[t!]
\begin{center}
\centerline{\includegraphics[width=0.88\hsize]{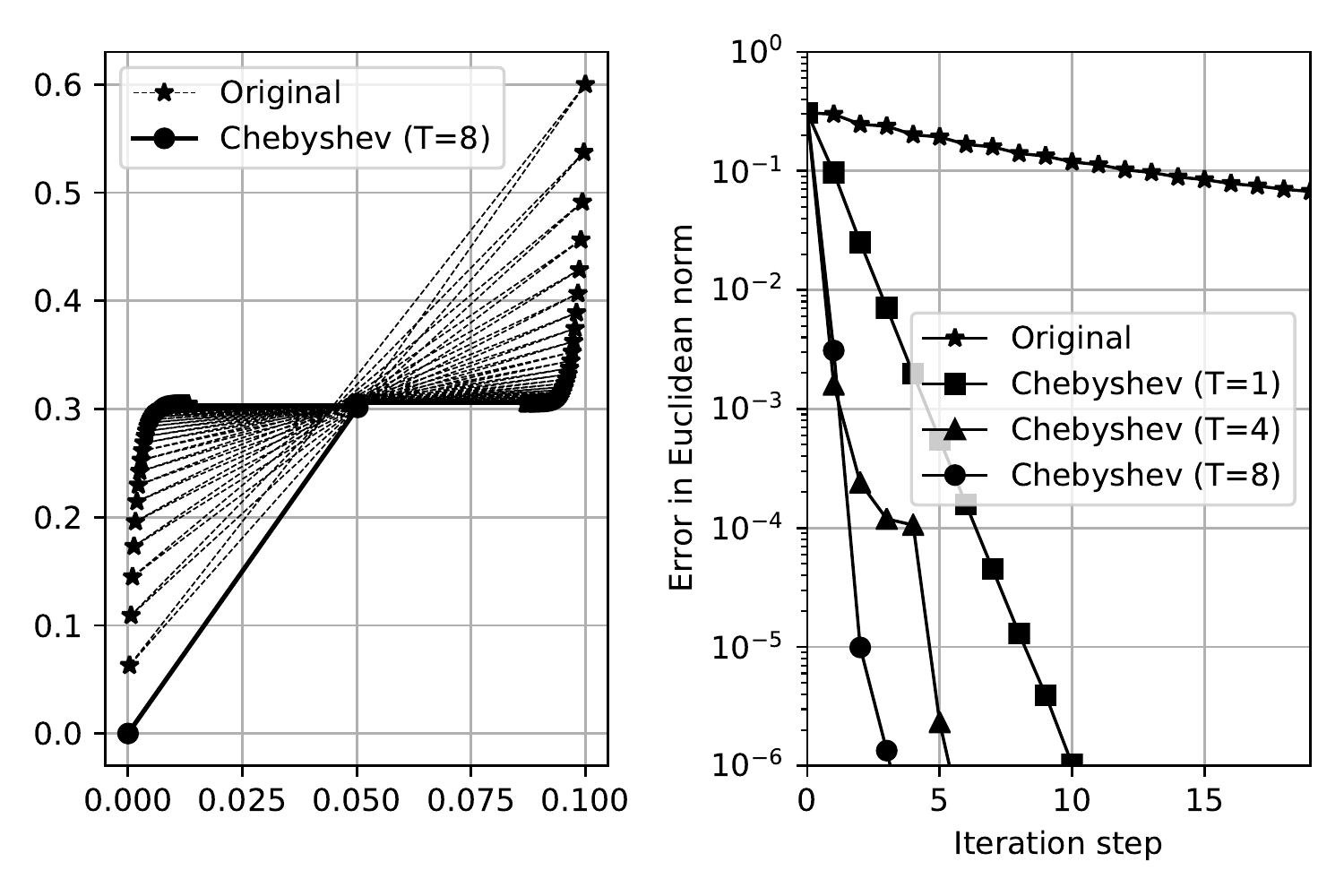}}
\caption{Solving the nonlinear equation: The solution of a nonlinear equation $\bm y = \bm x + \tanh (\bm x) (\bm y = (0.1, 0.6)^T)$ 
can be found by the fixed-point iteration $\bm x^{(k+1)} = \bm y - \tanh(\bm x^{(k)})$. 
The fixed point is $(0.0500, 0.3045)^T$.
The left figure indicates the search trajectories of the original fixed-point iteration and 
the Chebyshev-PSOR. 
The right figure presents the error from the fixed point. }
\label{nonlinear}
\end{center}
\vskip -0.2in
\end{figure}

If the initial point is sufficiently close to the solution, 
one can solve the nonlinear equation with a fixed-point iteration.
Figure \ref{nonlinear} illustrates the behaviors of two-dimensional fixed-point iterations for solving
a nonlinear equation
$
\bm y = \bm x + \tanh(\bm x).  	
$
We can see that the Chebyshev-PSOR yields  
much steeper error curves compared with the original iteration in Fig. \ref{nonlinear} (right).

Consider the nonlinear fixed-point iteration
	$\bm x^{(k+1)} = \tanh(\bm A \bm x^{(k)})$,			
where $\bm A \in \mathbb{R}^{512 \times 512}$.
The Jacobian $\bm J^*$ at the fixed point is identical 
to $\bm A$, and the minimum and maximum eigenvalues of $\bm J^*$ are $0$ and $0.9766$, respectively.
Figure \ref{theory} shows the normalized errors $\|\bm x^{(k)} - \bm x^*\|_2/n$ for the 
original and the Chebyshev-PSORs. This figure also includes 
the theoretical values $[\rho(\bm J^*)]^{2k}$ and 
$[q_{ch}(T)]^{2 k}$ 
for $a = 0.0234$ and $b = 1.0$.
We can observe that error curves of the Chebyshev-PSOR show a zig-zag shape as well.
The lower envelopes of the zig-zag lines (corresponding to periodically bounded errors)
and the theoretical values derived from Thm.~\ref{convergence_rate}
have almost identical slopes.

\begin{figure}[t!]
\begin{center}
\centerline{\includegraphics[width=0.88\hsize]{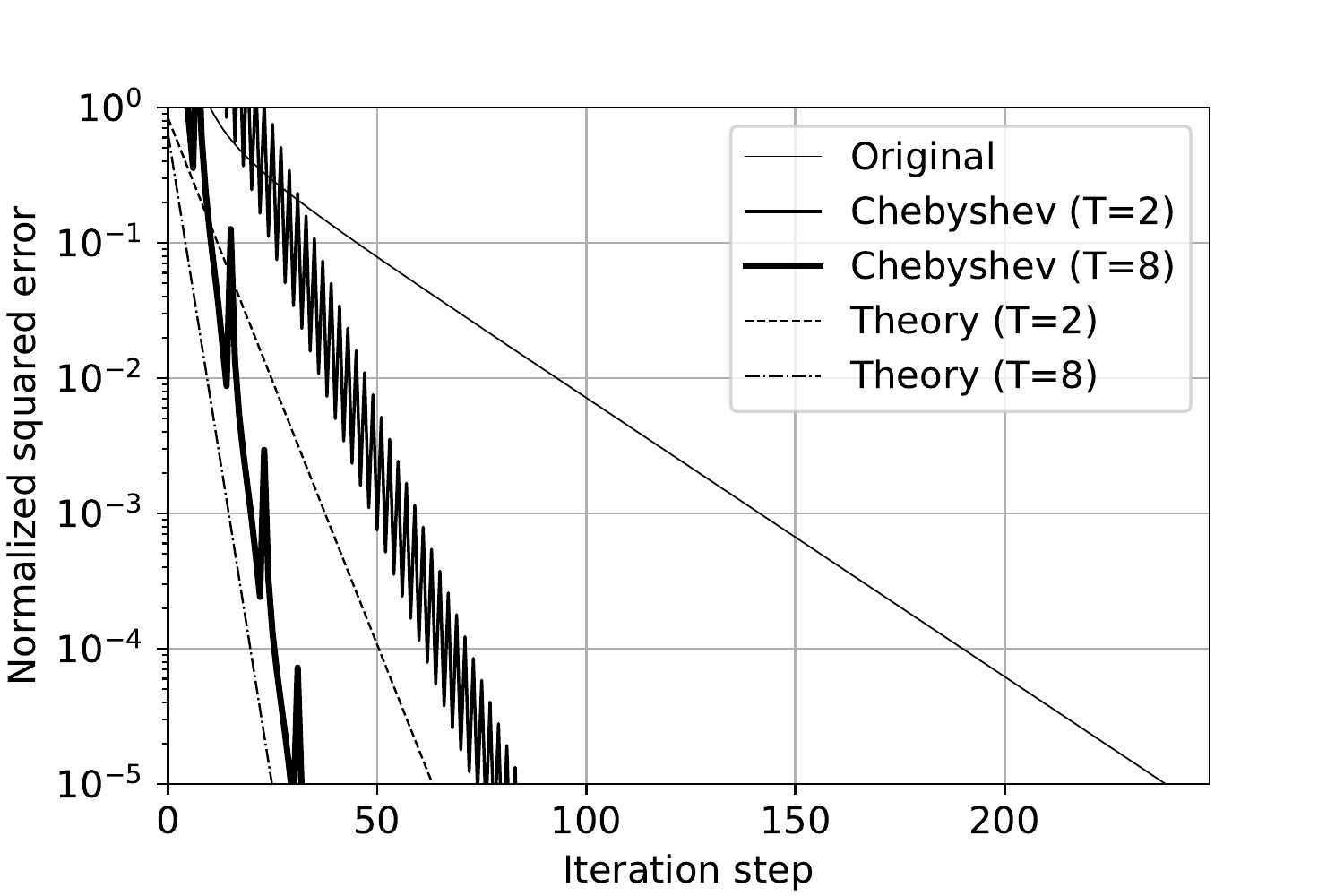}}
\caption{Nonlinear fixed-point iteration in a high-dimensional space: 
The fixed-point iteration is $\bm x^{(k+1)} = \tanh(\bm A \bm x^{(k)})$ and $\bm A$ is a random Gram matrix 
$\bm A = \bm M^T \bm M$ of size $512 \times 512$.
Each element of $\bm M$ follows an i.i.d. Gaussian distribution with zero mean and 
a standard deviation of 0.022. 
 The solid lines represent the normalized squared error $\|\bm x^{(k)} - \bm x^*\|^2_2/512$ for 
the original fixed-point iteration and the Chebyshev-PSORs $(T= 2, 8)$.
The fixed point is the origin. The dotted lines (Theory) indicate 
$[q_{CI}(T)]^{2k}$ where $T = 2, 8$, and $a = 0.0234, b = 1.0$.
}
\label{theory}
\end{center}
\vskip -0.2in
\end{figure}

\subsubsection{ISTA for sparse signal recovery}

ISTA \cite{Daubechies04}
 is a proximal gradient descent method designed for sparse signal recovery problems.
The setup of the sparse signal recovery problem discussed here is as follows:
Let $\bm x \in \mathbb{R}^n$ be a source sparse signal. We assume each element in $\bm x$ follows
an i.i.d. Bernoulli-Gaussian distribution; i.e., non-zero elements in $\bm x$ occur with probability $p$.
These non-zero elements follow the normal distribution ${\cal N}(0, 1)$.
Each element in a sensing matrix $\bm M \in \mathbb{R}^{m \times n}$ is assumed to be generated
randomly according to ${\cal N}(0, 1)$. The observation signal $\bm y$ is given by
$
	\bm y = \bm M \bm x + \bm w,	
$
where $\bm w \in \mathbb{R}^m$ is an i.i.d. noise vector whose elements follow ${\cal N}(0, \sigma^2)$.
Our task is to recover the original sparse signal $\bm x$ from the observation $\bm y$
as correctly as possible.
This problem setting is also known as compressed sensing \cite{donoho2006compressed, Candes06}.

A common approach for the sparse signal recovery problem described above is 
to use the Lasso formulation \cite{Tibshirani96, Efron04}:
\begin{equation}
\hat {\bm x} := \arg \min_{\bm x \in \mathbb{R}^n} \frac 1 2 \| \bm y - \bm M \bm x\|^2_2 + \lambda \|\bm x\|_1.		
\end{equation}

ISTA is a proximal gradient descent method {for solving Lasso problem.} 
The fixed-point iteration of ISTA is given by
\begin{eqnarray} \label{ISTAiteration1}
	\bm r_t &=& \bm s_t + \beta \bm M^T ( \bm y - \bm M \bm s_t), \\ \label{ISTAiteration2}
	\bm s_{t+1} &=& \eta(\bm r_t; \tau),
\end{eqnarray}
where $\eta(x; \tau)$ is the soft shrinkage function defined by
$
	\eta(x; \tau) := \text{sign}(x) \max \{|x| - \tau, 0 \}.		
$
Acceleration of the proximal gradient methods is an important topic for convex optimization~\cite{Bubeck15}.
FISTA \cite{Beck09} is an accelerated algorithm based on ISTA and Nesterov's acceleration~\cite{Nesterov83}.
FISTA and its variants \cite{Bioucas-Dias07}
achieve a much faster global convergence speed, $O(1/k^2)$, compared with that of the original ISTA with time complexity $O(1/k)$.

The softplus function $s_p(x)$ is defined as 
$
	s_p(x) := \frac{1}{\beta}\ln(1  + \exp(\beta x)).			
$
In the following experiments, we will use 
a {\em differentiable soft shrinkage function} defined by
\begin{equation}\label{eta}
	\tilde \eta(x; \tau) := s_p(x - \tau) + s_p(- (x + \tau))	
\end{equation}
with $\beta = 100$ instead of the soft shrinkage function $\eta(x; \tau)$ because
$\tilde \eta(x; \tau) $ is differentiable. 

Let $\bm G = \bm M^T \bm M$. The step-size parameter and shrinkage parameter are 
set as $\gamma = \tau = 1/\lambda_{max}(\bm G)$.
We are now ready to state 
the ISTA iteration (\ref{ISTAiteration1}), (\ref{ISTAiteration2}) in the form of the fixed-point iteration based on 
an affine composite update function:
\begin{equation}
	\bm x^{(k+1)} = \tilde \eta(\bm A \bm x^{(k)} + \bm b ; \tau),	
\end{equation}
where $\bm A := \bm I_n - \gamma \bm G$ and $\bm b := \gamma \bm M^T \bm y$.

\begin{figure}[t!]
\begin{center}
\centerline{\includegraphics[width=0.88\hsize]{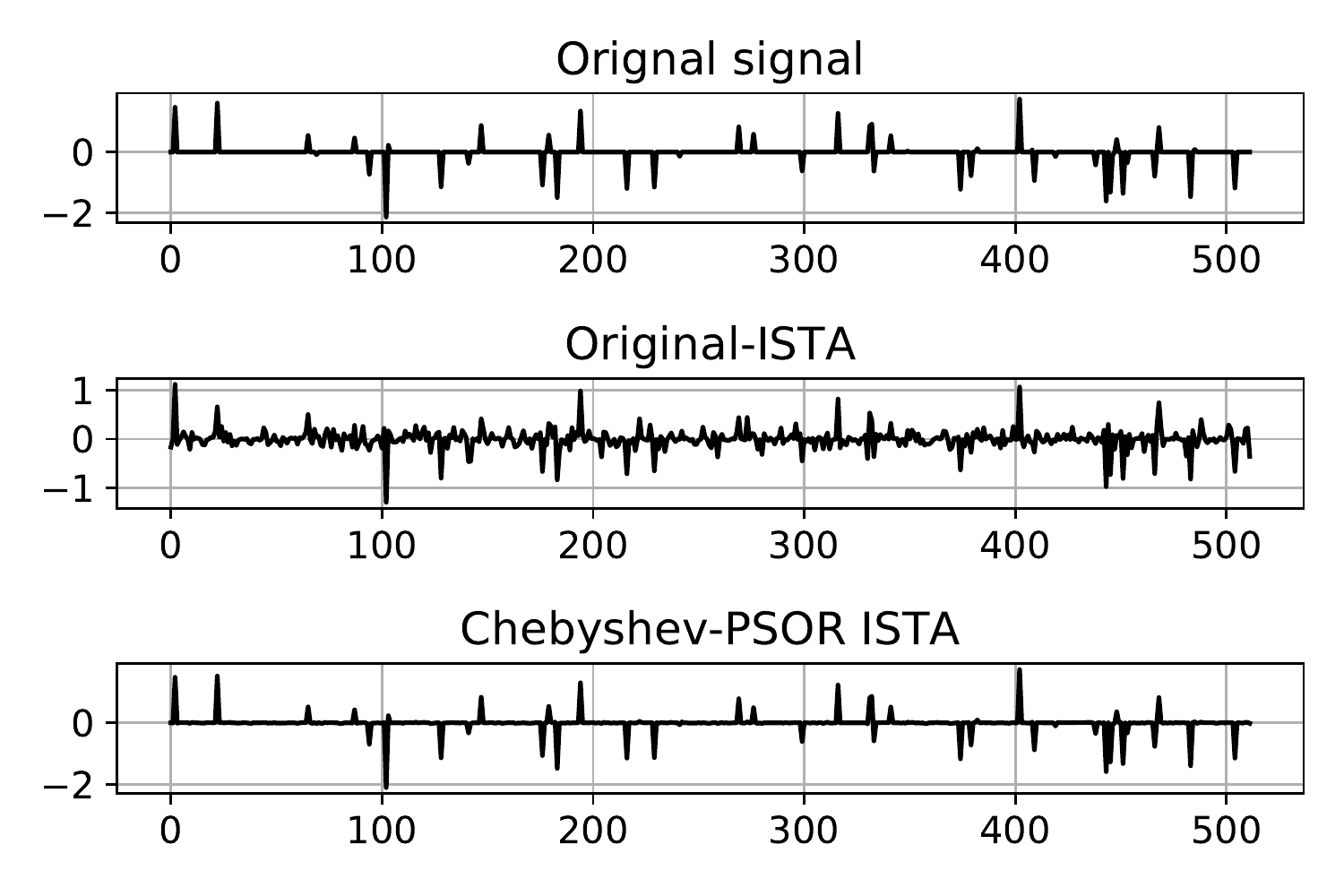}} 
\caption{Estimated signals : $n = 512, m = 256, p = 0.1, \sigma = 0.1, T = 8$. 
The middle and lower figures are estimated signals after 200 iterations.}
\label{ista_signal}
\end{center}
\vskip -0.2in
\end{figure}

\begin{figure}[t]
\begin{center}
\centerline{\includegraphics[width=0.88\hsize]{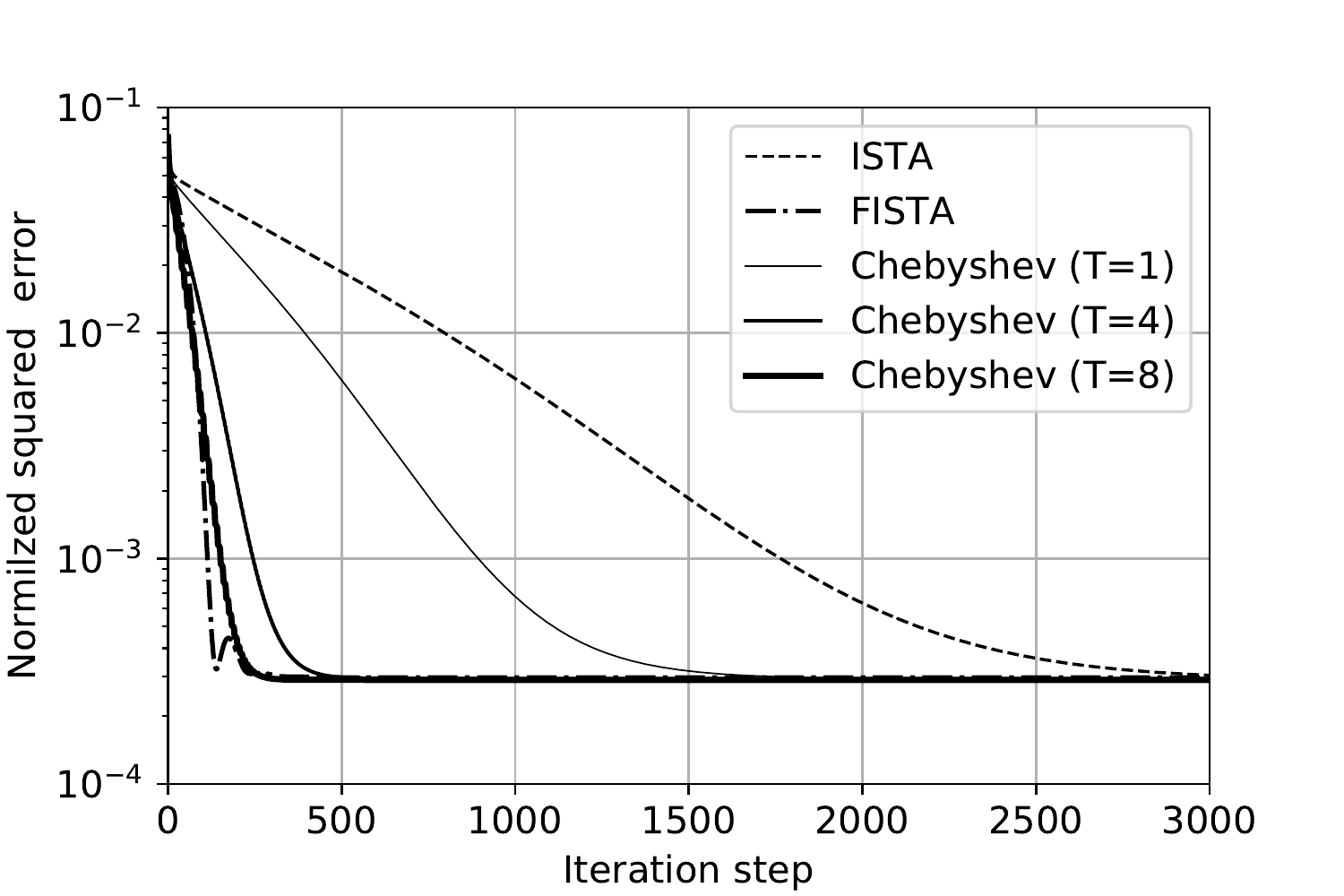}} 
\centerline{\includegraphics[width=0.88\hsize]{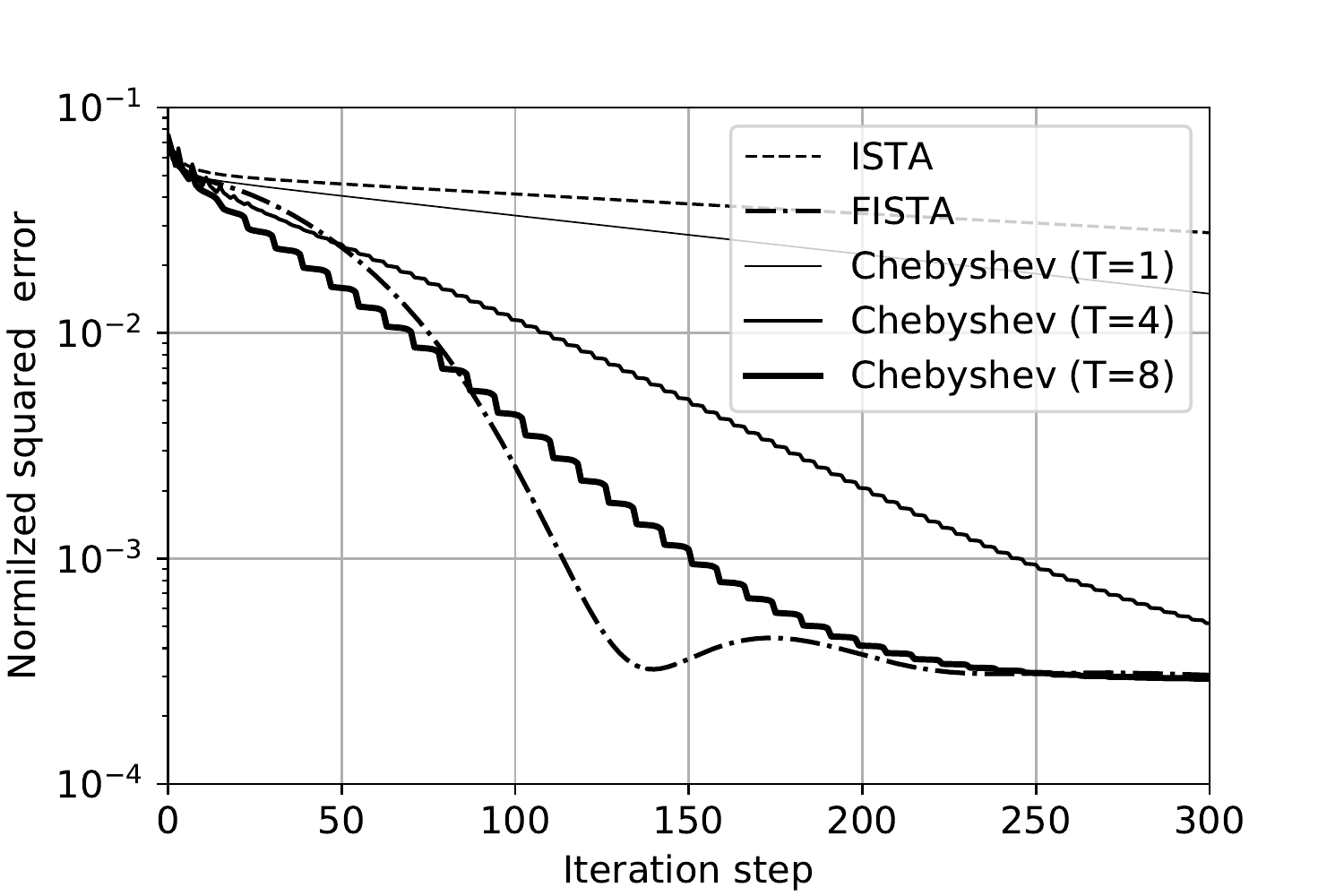}}
\caption{Normalized squared errors (upper: up to 3000 iterations, lower: up to 300 iterations): $n = 512, m = 256, p = 0.1, \sigma = 0.1$. 
The observation vector is modeled by $\bm y = \bm M \bm x + \bm w$, where $\bm M \in \mathbb{R}^{m \times n}$.
The normalized squared error $\|\bm x^{(k)} - \bm x\|^2/n$ is averaged over 1000 trials. As a reference, 
the error curves of FISTA \cite{Beck09} are also included.}
\label{ista_average}
\end{center}
\vskip -0.2in
\end{figure}

\begin{figure}[t]
\begin{center}
\centerline{\includegraphics[width=0.88\hsize]{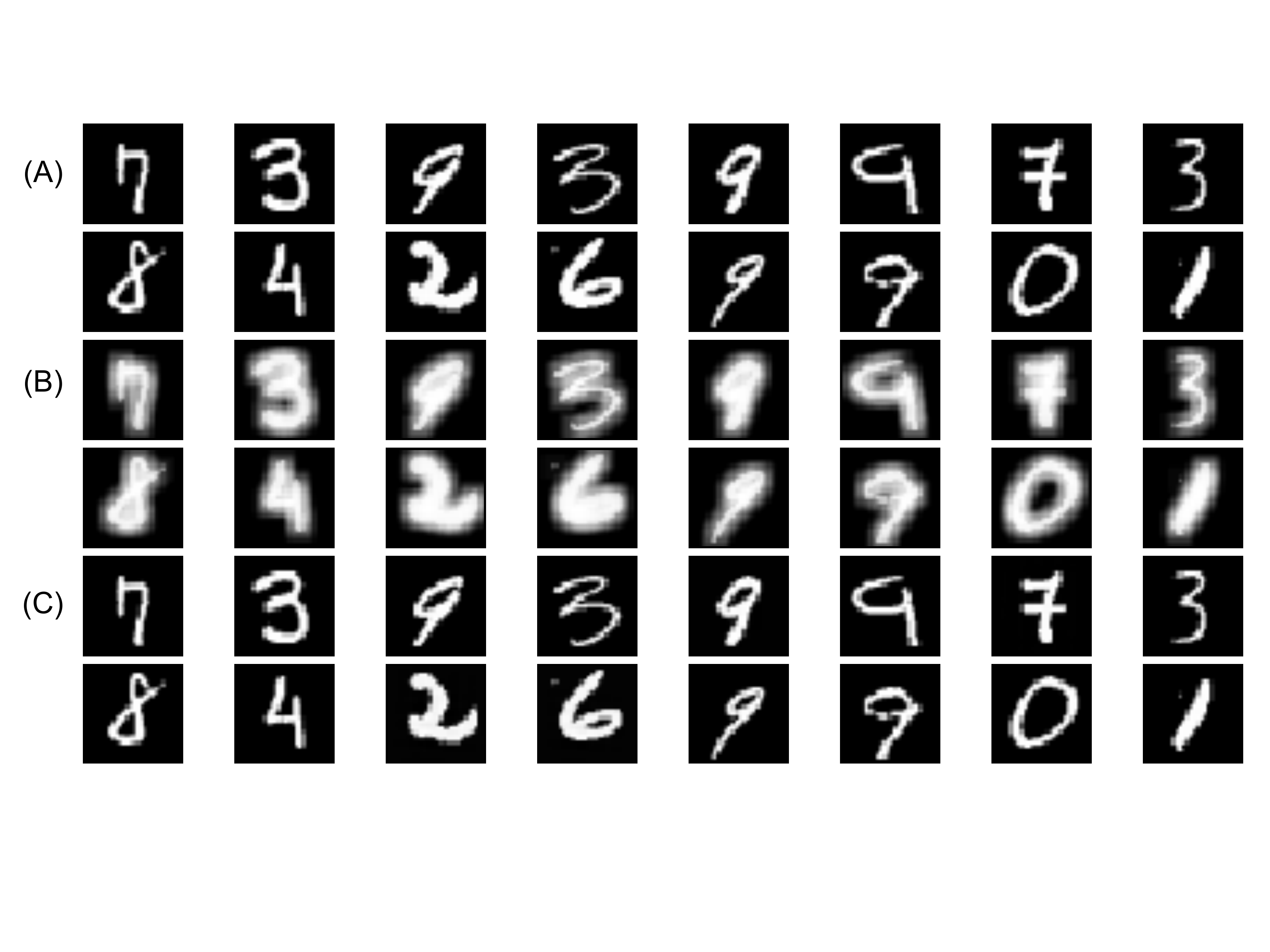}}
\vspace{-1.0cm}
\centerline{\includegraphics[width=0.88\hsize]{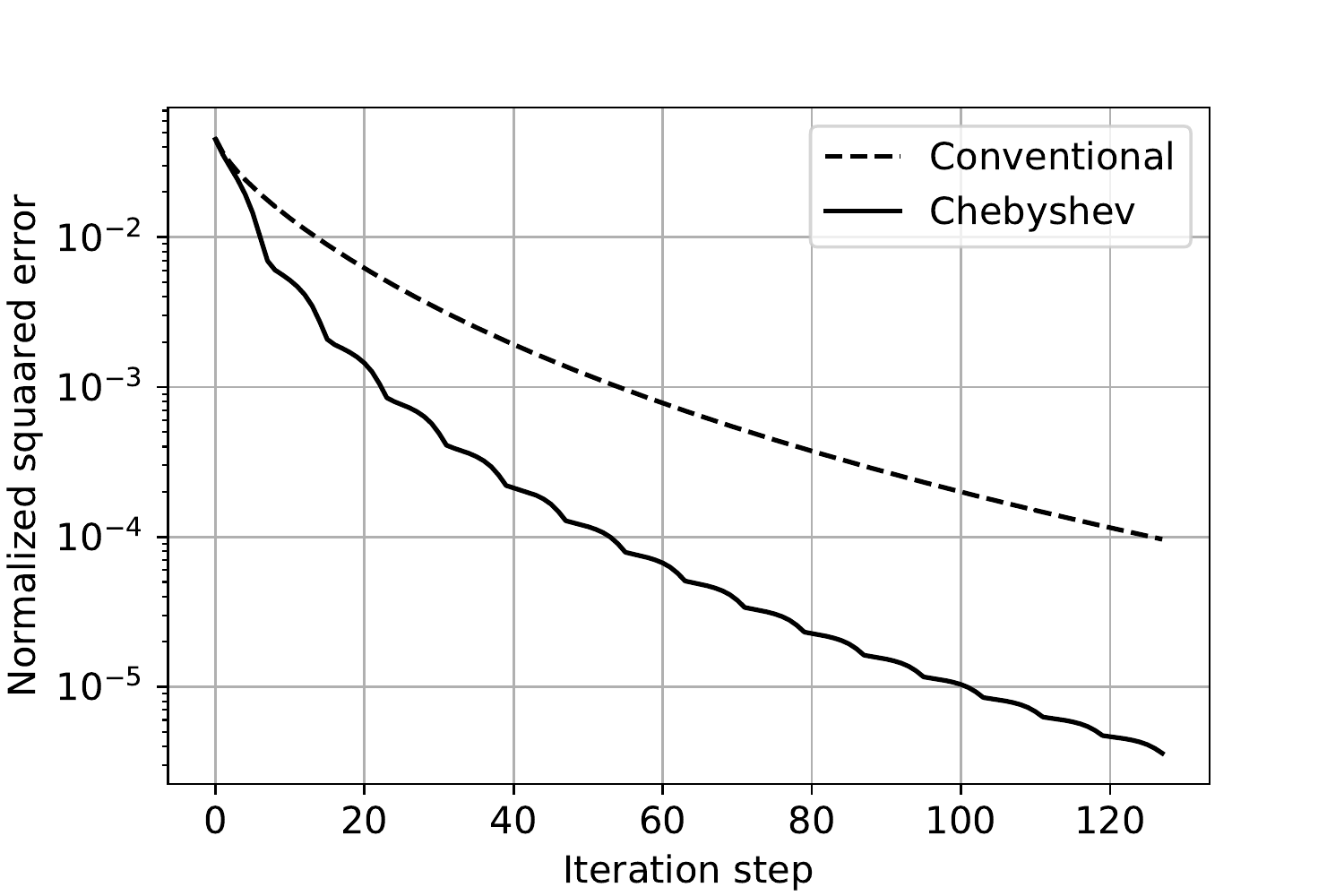}}
\caption{Nonlinear deblurring by modified Richardson iteration: 
  (A) Original image, (B) blurred image, and (C) recovered image by Richardson iteration with Chebyshev-PSOR.
  The blurring process is as follows: A $7 \times 7$ convolutional kernel with the value 0.1 except for the center (the value 1.5) 
 is first applied to a normalized MNIST image in $[0,1]^{28 \times 28}$, and then, a sigmoid function is applied to the convoluted image. The parameter $\omega$ is set to $0.8$.
The number of iterations is 128, $T=8$, $\lambda_{min}(\bm B) = 0.18, \lambda_{max}(\bm B) = 0.98$.
The graph represents normalized errors $\|\bm x^{(k)} - \bm x\|^2_2/28^2$. }
\label{blur}
\end{center}
\vskip -0.2in
\end{figure}

Figure \ref{ista_average} presents estimated signals (ISTA and Chebyshev-PSOR ISTA) after 200 iterations.
Because the original ISTA is too slow to converge, 200 iterations are not enough to recover the original signal 
with certain accuracy. In contrast, Chebyshev-PSOR ISTA 

Figure \ref{ista_average} indicates the averaged normalized squared error for 1000 trials.
The upper and lower figures in Fig. \ref{ista_average} correspond the maximum number of 
iterations 3000 and 300, respectively.  
The Chebyshev-PSOR ISTA ($T=8$) requires only 250--300 iterations 
to reach the point achieved by the original ISTA after 3000 iterations.
The convergence speed of the Chebyshev-PSOR 
ISTA ($T=8$) is similar to that of FISTA but 
the proposed scheme provides smaller squared errors 
when the number of iterations is less than 70.
This result implies that the proposed scheme can be used 
for a wide range of convex and non-convex problems 
as an alternative to FISTA and Nesterov's acceleration.

\subsubsection{Modified Richardson iteration for deblurring}

Assume that a signal $\bm x \in \mathbb{R}^n$ is distorted 
by a nonlinear process $\bm y = f(\bm x)$.
The modified Richardson iteration \cite{saad2003iterative} is usually known as a method for
solving linear equations, but it can be applied to solve a nonlinear equation.
The fixed-point iteration is given by 
$
	\bm x^{(k+1)} = \bm x^{(k)} + \omega ( \bm y - f(\bm x^{(k)} )  )	
$
where $\omega$ is a positive constant. 
Figure \ref{blur} shows the case where $f$ expresses a nonlinear blurring process to the MNIST images.
We can also discern that the Chebyshev-PSOR can accelerate convergence.

\section{Conclusion}\label{sec_con}
In this study, we first present a plausible theoretical interpretation of convergence acceleration of DUGD by introducing Chebyshev steps and then demonstrate novel accelerated fixed-point iterations called Chebyshev-PSOR as a promising 
application of Chebyshev steps.

In the first half of the study, 
{we introduce Chebyshev steps as a step-size sequence of GD.  We show}
that Chebyshev steps reduce the convergence rate compared with a naive GD 
and the convergence rate approaches the lower bound of the first-order method {as $T$ increases.}
Numerical results show that Chebyshev steps well explain the {zig-zag shape of}  learned step-size parameters of DUGD, which can be considered as a plausible interpretation of the learned step-size.
This finding indicates the promising potential of deep unfolding for promoting 
convergence acceleration and
suggest that trying to interpret the learned parameters in deep-unfolded 
algorithms would be fruitful from the theoretical viewpoint.


In the second half of the paper, based on the above analysis, we introduce 
a novel accelerated fixed-point iteration called Chebyshev-PSOR.
Theoretical analysis 
revealed the local convergence behavior of the Chebyshev-PSOR, {which is supported by experimental results.}
The Chebyshev-PSOR can be applied to a wide range of iterative algorithms with fixed-point iterations, 
{and} requires no training process to adjust the PSOR factors, {which is 
preferable to deep unfolding.
{It is also noted that Chebyshev-PSOR is applicable to the Landweber algorithm in \emph{infinite-dimensional} Hilbert space, which is omitted due to page limit (see~\cite{wadayama2020chebyshev} for details).} 

In the numerical experiments, we naively assumed that the maximum and minimum eigenvalues for Chebyshev steps were known. 
In practice, however, the computational complexity of eigenvalue computation is a non-negligible problem.
We can roughly estimate the eigenvalues using, e.g., the power method, eigenvalue inequalities like Gershgorin circle theorem, and random matrix theory such as {the Marchenko-Pastur distribution law and} the semi-circle law. 
It will be important to study such a preprocessing technique for practical purposes.


 \section*{Acknowledgement}
This study is partly supported by JSPS Grant-in-Aid for Scientific Research (B) Grant Number 19H02138  (TW)  and for Early-Career Scientists Grant Number 19K14613 (ST), and the Telecommunications Advancement Foundation (ST).  


\bibliographystyle{IEEEtran}
\bibliography{IEEEabrv,Cheb}

\newpage
\setcounter{page}{1}
\section*{Supplemental Material for \newline ``Convergence Acceleration via Chebyshev Step:  Plausible Interpretation of Deep-Unfolded Gradient Descent,'' \newline Satoshi Takabe and Tadashi Wadayama} 
\appendices

\section{Chebyshev polynomials}\label{sec_poly}

The analysis in this paper is based on the properties of Chebyshev polynomials. 
{In this subsection, we review some important properties of Chebyshev polynomials~\cite{Mason03}.}

The Chebyshev polynomial $C_n(x)$ of degree $n$ is defined using the following three-term recurrence relation: 
\begin{equation}
C_{n+1}(x)=2x C_{n}(x)-C_{n-1}(x),\label{eq_ch0}
\end{equation}
 where $C_{0}(x)=1$ and $C_{1}(x)=x$.
For example, we have $C_2(x)=2x^2-1$ and $C_3(x)=4x^3-3x$.
{Chebyshev polynomials $C_i(x) (i = 1,2,3,4)$ are depicted in Fig. \ref{cheb_poly}. The Chebyshev polynomials are also 
characterized by the following identity:}
\begin{equation}
C_n(x) = \begin{cases}
\cos(n\cos^{-1}(x)) & (|x|\le 1)\\
\cosh(n\cosh^{-1}(x))  & (|x|\ge 1)
\end{cases}.\label{eq_ch1}
\end{equation}
We then get $|C_n(x)|\le 1$ for any $x\in[-1,1]$.
For $|x|\ge 1$, we have another representation given as
\begin{equation}
C_n(x)=\frac{(x+\sqrt{x^2-1})^n+(x-\sqrt{x^2-1})^n}{2}. \label{eq_ch2}
\end{equation}

{From the recursion (\ref{eq_ch0}), it is immediately observed that }
$C_n(x)$ is a polynomial of order $n$ whose highest coefficient is $2^{n-1}$.
As suggested by (\ref{eq_ch1}), the zeros of $C_n(x)$ are given by 
\begin{equation}
x_k=\cos\left(\frac{2k+1}{2n}\pi\right).\quad (k=0,\dots,n-1) \label{eq_ch3}
\end{equation}
{These zeros are called {\em Chebyshev nodes}, which are 
commonly used for polynomial interpolation.}

Chebyshev polynomials have extremum points in the interval $[-1,1]$, which is given as follows
\begin{equation}
\tilde x_k=\cos\left(\frac{k\pi}{n}\right), \quad (k=0,\dots,n-1) \label{eq_ch4}
\end{equation}
including both end points.
Their absolute values equal to $1$, and their signs change by turns. 
This equioscillation {property} leads to the so-called minimax property. 
Let $C[a,b]$ be the Banach space defined by $\ell_\infty$-norm
$\|f\|_\infty := \max_{x\in[a,b]} |f(x)|$.
Then, $C[a,b]$ is a space of continuous polynomials on $[a,b]$.
Chebyshev polynomials minimize the $\ell_\infty$-norm as follows.
\begin{theorem}[\cite{Mason03}, Col. 3.4B]\label{thm_ch} 
In $C[-1,1]$, 
$2^{1-n}C_n(x)$ is a monic polynomial of degree $n$ having the smallest norm 
$\|2^{1-n}C_n(x)\|_\infty=2^{1-n}$.
\end{theorem}

\begin{figure}[t]
   \centering
   \includegraphics[width=0.88\hsize]{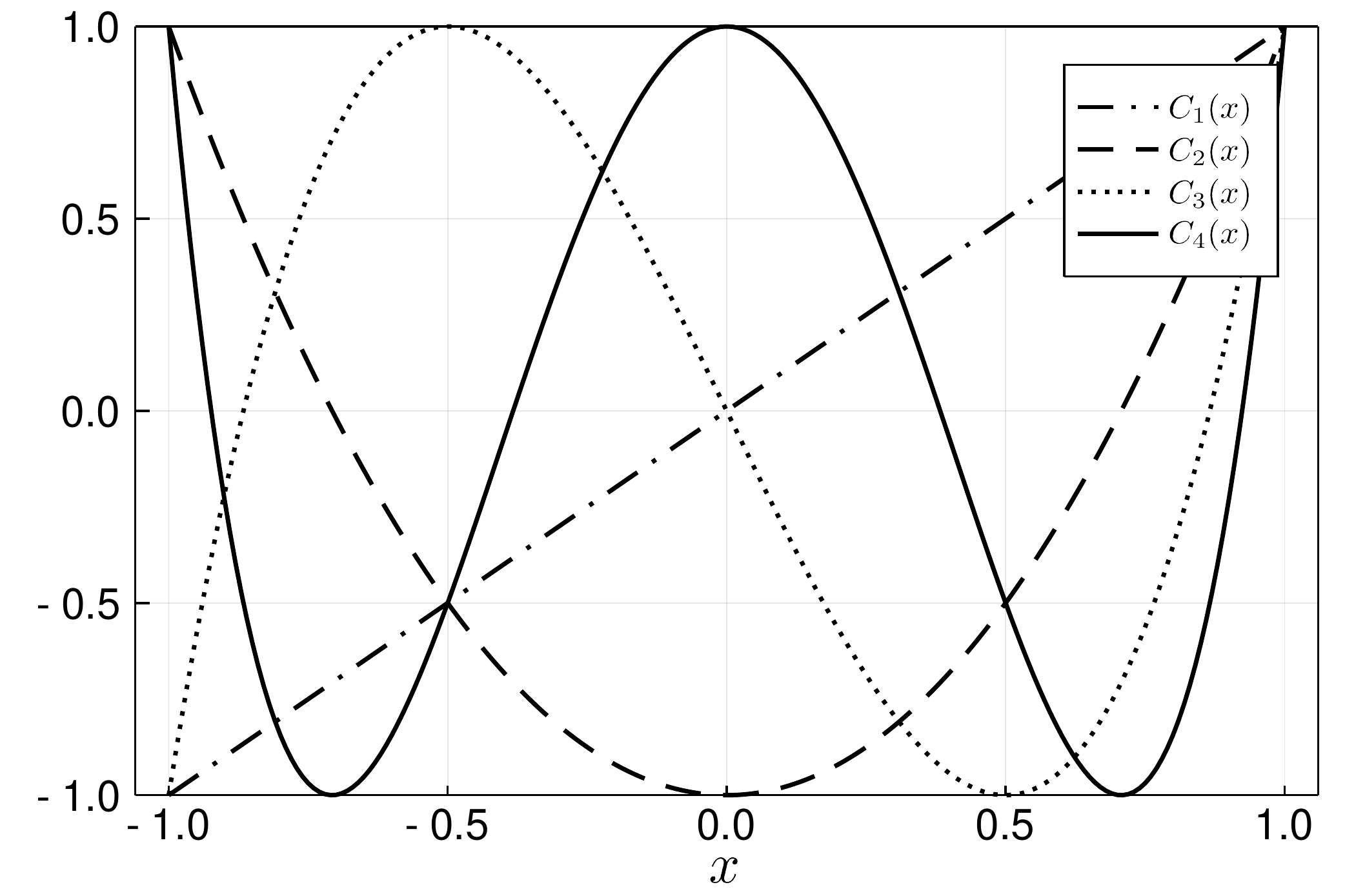}
    \caption{Chebyshev polynomials $C_1(x), C_2(x), C_3(x), C_4(x)$.}
    \label{cheb_poly}
\end{figure}

\section{Proof of Theorem~\ref{thm_step}}\label{app_1}

We first show the following lemma.
\begin{lemma}\label{lem_spec}
Suppose that $b>a>0$.
Let $D\subset C[a,b]$ be a subspace of polynomials of $z$ on $[a,b]$ represented by 
$\prod_{k=0}^{n-1}(1-\alpha_k z)$ for any $\alpha_0,\cdots,\alpha_{n-1}\in \mathbb{R}$.
We define the normalized Chebyshev polynomial $\hat\varphi(z)$ of degree $n$ as
\begin{equation}
\hat \varphi(z) 
:=  \frac{C_n\left(\frac{2z-a-b}{b-a}\right)}{C_n\left(-\frac{a+b}{b-a}\right)}. \label{eq_norm_ch}
\end{equation}
Then, the following statements hold.

(a) The function $\hat \varphi:[a,b]\rightarrow \mathbb{R}$ belongs to $D$ as a result of setting $\alpha_k=(\gamma_k^{ch})^{-1}$ ($k=0,1,\dots,n-1$).

(b) The function $\hat \varphi:[a,b]\rightarrow \mathbb{R}$ is a polynomial in $D$ that minimizes norm $\|\cdot\|_\infty$.
\end{lemma}

\begin{IEEEproof}
(a) Using Chebyshev nodes $\{x_k\}_{k=0}^{T-1}$ given as (\ref{eq_ch3}), we get $C_n(x)= \prod_{k=0}^{n-1}(x-{x}_k)$.
Then, by the affine transformation from $[a,b]$ to $[-1,1]$, we have
\begin{equation}
C_n\left(\frac{2z-a-b}{b-a}\right) =\prod_{k=0}^{n-1}\left(z-\frac{1}{\gamma_k^{ch}}\right). \label{eq_cn_zero}
\end{equation}
Because $C_n(-(a+b)/(b-a)) =\prod_{k=0}^{n-1}[-(\gamma_k^{ch})^{-1}]$,
we have $\hat \varphi(z) = \prod_{k=0}^{n-1}(1-\gamma_k^{ch} z)$, which indicates that the statement holds.

(b) We show that $\hat\varphi(z)$ is a minimizer of $\|\cdot\|_\infty$ among functions in $D$ by indirect proof.
Assume that there exists $\tau(z)\in D$ of {at most} degree $n$, except for $\hat\varphi(x)$ satisfying $\|\hat\varphi(z)\|_\infty>\|\tau(z)\|_\infty$.
Using $\tilde x_k$ ($k=0,1,\dots,n$) given by (\ref{eq_ch4}), and
$\tilde z_k := (a+b)/2 + (b-a)\tilde x_k/2(\in [a,b])$ are extremal points of $\hat\varphi(z)$ including both ends.
Particularly, the sign of the extremal value at $\tilde x_k$ (or $\tilde z_k$) changes alternatively; 
$C_n(\tilde x_n)=1$, $C_n(\tilde x_{n-1})=-1$, $C_n(\tilde x_{n-2})=1$, and so on (or 
$\hat\varphi(\tilde z_n)=\varphi_0$, $\hat\varphi(\tilde z_{n-1})=-\varphi_0$, $\hat\varphi(\tilde z_{n-2})=\varphi_0$, and so on when $\varphi_0:=1/C_n(-(a+b)/(b-a))$) hold~\cite[Lemma 3.6]{Mason03}.
The assumption indicates that $n+1$ inequalities, $\tau(\tilde z_{n})< \varphi_0$, $\tau(\tilde z_{n-1})> -\varphi_0$, $\tau(\tilde z_{n-2})< \varphi_0$, and so on hold; that is, a polynomial $\delta(z):=\tau(z)-\hat\varphi(z)$ of degree of at most $n$ has $n$ zeros in $[a,b]$.

However, because $\tau(z), \hat\varphi(z)\in D$, the constant term of $\delta(z)$ is equal to zero, which suggests that $\delta(z)$ has at most $n-1$ zeros in $[a,b]$.
This results in a contradiction of the assumption and shows that $\hat \varphi:[a,b]\rightarrow \mathbb{R}$ minimizes the norm $\|\cdot\|_\infty$ in $D$.  
\end{IEEEproof}

Thus, it is straightforward to prove Thm.~\ref{thm_step} from this lemma.  
\begin{IEEEproof}[proof of Thm.~\ref{thm_step}]
Using the notation of Lem.~\ref{lem_spec}, we notice that $a=\lambda_1$, $b=\lambda_n$, and
\begin{align}
 \rho^{\mathrm{upp}}( \bm Q^{(T)}) &= \max_{ \lambda\in [\lambda_1,\lambda_n] }|\beta_T(\lambda)| \nonumber\\
&= \left\|\beta_T(\lambda) \right\|_\infty.
\end{align}
From Lem.~\ref{lem_spec}, the Chebyshev steps of length $T$ form a sequence that minimizes $\rho^{\mathrm{upp}}( \bm Q^{(T)})$. 
\end{IEEEproof}

\section{Proof of Lemma~\ref{lem_qt}}\label{app_rate}
Let $\beta^{ch}_T(\lambda)$ be the function $\beta_T(\lambda)$ with Chebyshev steps, which is given as
$\beta_T^{ch}(\lambda):=\prod_{t=0}^{T-1}(1-\gamma_t^{ch}\lambda)$.
The spectral radius when the Chebyshev steps are applied is given as follows:
\begin{align}
&\rho^{\mathrm{upp}}(\bm  Q^{(T)}_{\mathrm{ch}})\nonumber\\
&= \max_{\lambda\in[\lambda_1,\lambda_n]}\left|\beta^{ch}_T(\lambda)\right| \nonumber\\
&= \max_{\lambda\in[\lambda_1,\lambda_n]}
\left|C_T\left(-\frac{\kappa+1}{\kappa-1}\right)^{-1}
C_T\left(\frac{2\lambda-\lambda_n-\lambda_1}{\lambda_n-\lambda_1}\right) \right| \nonumber\\
&= \left|C_T\left(-\frac{\kappa+1}{\kappa-1}\right)\right|^{-1} \nonumber\\
&= \left\{\frac{1}{2}\left[ \left(\frac{\sqrt\kappa+1}{\sqrt\kappa-1}\right)^T
+\left(\frac{\sqrt\kappa-1}{\sqrt\kappa+1}\right)^T\right]\right\}^{-1}\nonumber\\
&= \mathrm {sech}\left[T \cosh^{-1}\left(\frac{\kappa+1}{\kappa-1}\right) \right] , \label{eq_qt}
\end{align}
where $\kappa:=\kappa(\bm A)$.
The third line is derived from Lem.~\ref{lem_spec} in App.~\ref{app_1}.
We use (\ref{eq_ch2}) to obtain the fifth line and (\ref{eq_ch1}) for the sixth line.
\hfill \IEEEQEDclosed

\section{Proof of Theorem~\ref{thm_main}}\label{app_2}

From (\ref{eq_qt}), we get
\begin{align}
&\rho^{\mathrm{upp}}(\bm  Q^{(T)}_{\mathrm{ch}})^{-1}-\rho( \bm Q^{(T)}_{\mathrm{const}})^{-1}\nonumber\\
& =\frac{(\sqrt\kappa+1)^{2T}+(\sqrt\kappa-1)^{2T}-2(\kappa+1)^{T}}{2(\kappa-1)^{T}}.
\label{eq_in11}
\end{align}
If we set $X:=\sqrt \kappa (>1)$, the $(2t)$th coefficient of $(X+1)^{2T}/2+(X-1)^{2T}/2- (X^2+1)^T$
is given by $\binom{2T}{2t}-\binom{T}{t}$, and $(2t+1)$th coefficients are equal to zero. 
From the Vandermonde identity
\begin{equation}
\binom{m+n}{r} = \sum_{k=0}^{r}\binom{m}{k}\binom{n}{r-k},
\end{equation}
we find 
\begin{equation}
\binom{2T}{2t} = \sum_{l=0}^{2t}\binom{T}{l}\binom{T}{2t-l}\ge \binom{T}{t}^2\ge \binom{T}{t},
\end{equation}
(equality holds only when $t=0,T$). 
This indicates that $\rho^{\mathrm{upp}}( \bm Q^{(T)}_{\mathrm{ch}})^{-1}>\rho( \bm Q^{(T)}_{\mathrm{const}})^{-1}$ holds,
which leads to (\ref{eq_ch_rho}) using $\rho( \bm Q^{(T)}_{\mathrm{ch}})<\rho^{\mathrm{upp}}(\bm  Q^{(T)}_{\mathrm{ch}})$. 
\hfill \IEEEQEDclosed

\section{Spectral radius and loss minimization in DUGD}\label{sec_ms}

Here, we study the relationship between the MSE and spectral radius $\rho( \bm Q^{(T)})$
that bridges the gap between Chebyshev steps and learned step-size sequences in DUGD.
The training process of deep unfolding consists of minimizing a loss function 
{regarding the MSE $L( \bm x^{(T)})=\| \bm x^{(T)}-\bm {x}^{\ast} \|_2^2/n$ between a tentative estimate and the solution.
We show that minimizing the MSE loss also reduces the spectral radius 
$\rho( \bm Q^{(T)})$. This result justifies the use of the deep unfolding approach 
for accelerating the convergence speed.}

Let $p(\bm {x}^{(0)})$ be an isotropic probability density function (PDF)~\footnote{{A PDF $p(\bm {x})$ is called isotropic if there exists a PDF $p_0(\bm x)$ satisfying $p(\bm x) = p_0(\|\bm x\|_2)$}.} satisfying $0<\mathsf{E}_{ \bm x^{(0)}}\|\bm  x^{(0)}\|_2^2<\infty$.
We define $\bm {x}^{(0)}\in\mathbb{C}^n$ as a random initial point over $p( \bm x^{(0)})$.
The Hermitian matrix  $\bm A$ {can be }decomposed by $ \bm A =  \bm U \bm \Lambda  \bm U^{\rm H}$ using unitary matrix $ \bm U$ and diagonal matrix $ \bm \Lambda:=\mathrm{diag}(\lambda_1,\dots,\lambda_n)$.
Then, we have 
\begin{align}
\mathsf{E}_{ \bm x^{(0)}}L( \bm x^{(T)})
&= \frac{1}{n} \mathsf{E}_{ \bm x^{(0)}}\left\|\prod_{t=0}^{T-1}( \bm I_n-\gamma_t  \bm U \bm \Lambda  \bm U^{\rm H}) \bm  x^{(0)}\right\|_2^2\nonumber\\
&= \frac{1}{n}  \mathsf{E}_{ \bm x^{(0)}}\left\| \bm U\left(\prod_{t=0}^{T-1}(\bm  I_n-\gamma_t  \bm \Lambda )\right)  \bm U^{\rm H} \bm x^{(0)}\right\|_2^2\nonumber\\
&= \frac{1}{n}  \mathsf{E}_{\bm  x^{(0)}}\left\| \left(\prod_{t=0}^{T-1}(\bm  I_n-\gamma_t  \bm \Lambda )\right)  \bm U^{\rm H} \bm x^{(0)}\right\|_2^2\nonumber\\
&:=  \frac{1}{n}  \mathsf{E}_{\bm  x^{(0)}}\left\|  \bm D^{(T)}  \bm U^{\rm H} \bm x^{(0)}\right\|_2^2
\end{align}
where $\bm {D}^{(T)}:=\mathrm{diag}(\beta_T(\lambda_i))$ is the diagonal matrix whose diagonal elements are eigenvalues of $\bm {Q}^{(T)}$.
Introducing column vectors of $ \bm U$ by $\bm {U} := (\bm u_1, \dots, \bm u_n)$
and $j:=\arg \max_{i}|\beta_T(\lambda_i)|$, we get
\begin{align}
\mathsf{E}_{\bm x^{(0)}}L(\bm x^{(T)})
&=  \frac{1}{n}  \mathsf{E}_{\bm x^{(0)}}\left\|\sum_{i=1}^n \bm D^{(T)}_{i,i}  \bm u_i^{\rm H} \bm x^{(0)}\right\|_2^2\nonumber\\
&\ge \frac{1}{n} \left|\beta_T(\lambda_j) \right|^2
\mathsf{E}_{\bm x^{(0)}}\left\|\bm u_j^{\rm H}\bm x^{(0)}\right\|_2^2\nonumber\\
&:= \frac{C'}{n}  \left|\beta_T(\lambda_j) \right|^2,
\end{align}
where $C':= \mathsf{E}_{\bm x^{(0)}}\|\bm u_j^{\rm H} \bm x^{(0)}\|^2$ is a positive constant because the PDF of $\bm {x}^{(0)}$ is assumed to be isotropic.
Recalling that $\rho(\bm Q^{(T)})=\max_{i}\left|\beta_T(\lambda_i)\right|=\left|\beta_T(\lambda_j) \right|$,
we have the following theorem.

\begin{theorem}\label{thm_spec}
Let $\bm {x}^{(0)}$ be a random vector over an isotropic PDF $p(\bm {x}^{(0)})$ satisfying $0<\mathsf{E}_{\bm  x^{(0)}}\| \bm x^{(0)}\|_2^2<\infty$.
Then, for any $T\in\mathbb{N}$, there exists a positive constant $C$ satisfying
\begin{equation}
{\rho( \bm Q^{(T)})}
\le C \sqrt{n\mathsf{E}_{\bm  x^{(0)}}L( \bm x^{(T)})}.
\label{eq_mse}\end{equation}
\end{theorem}

This theorem claims that minimizing the MSE loss function in DUGD reduces
 the corresponding spectral radius of $ \bm Q^{(T)}$, thus implying that appropriately learned step-size parameters can accelerate the convergence speed of DUGD.

\begin{figure}[t]
   \centering
   \includegraphics[width=0.88\hsize]{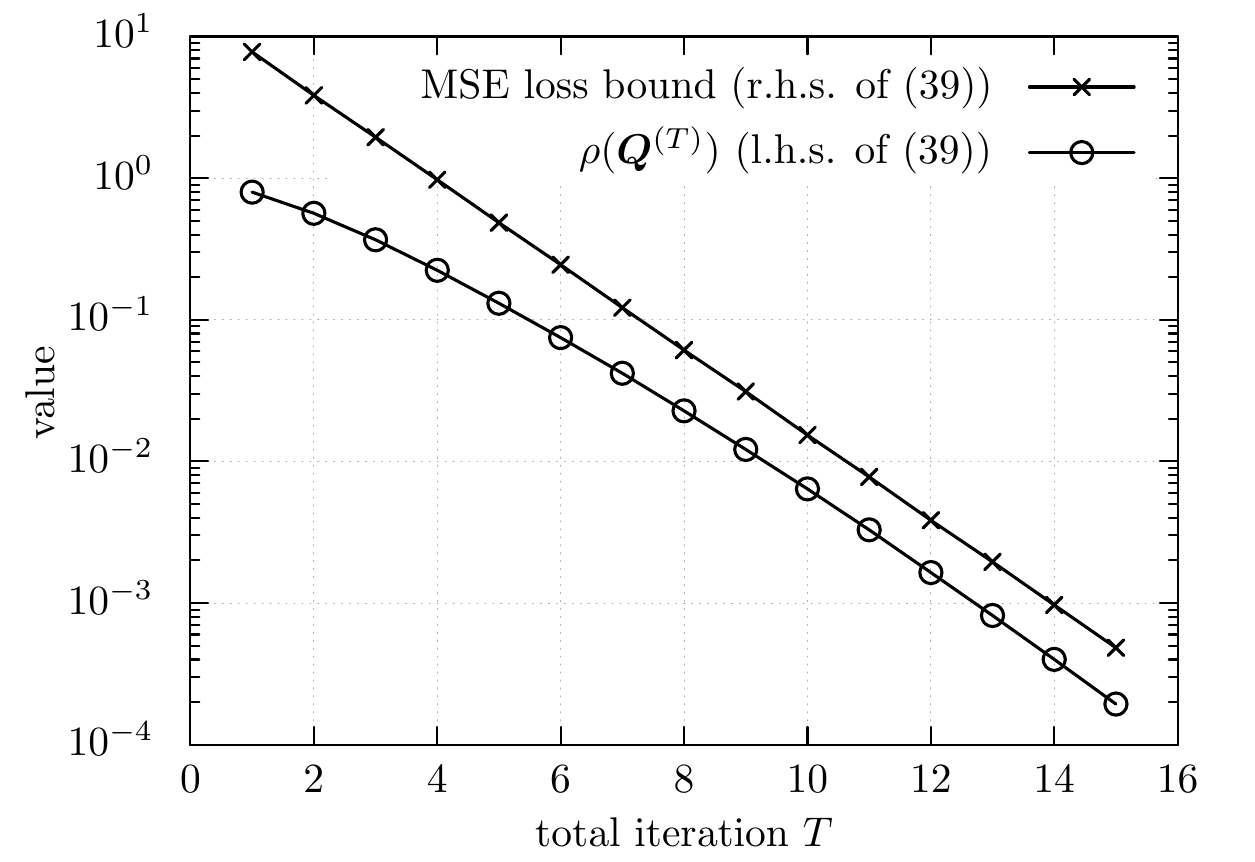}
    \caption{Comparison of the MSE loss bound in Thm.~\ref{thm_spec} and spectral radius $\rho( \bm Q^{(T)})$ in DUGD
     when $(n,m)=(300,1200)$. 
}
    \label{fig_loss}
    \vskip -0.2in
\end{figure}

We numerically verified the relation between the MSE loss in DUGD and the corresponding spectral radius $\rho( \bm Q^{(T)})$ {in Thm.~\ref{thm_spec}} up to $T\!=\!15$ of incremental training.
Figure~\ref{fig_loss} shows an example of the comparison when $(n,m)\!=\!(300,1200)$. 
The experimental conditions follow \label{sec_nc}.
To estimate the average MSE loss $\mathsf{E}_{ \bm x^{(0)}}L(\bm {x}^{(T)})$, we used empirical MSE loss after the $T$th generation. 
The constant $C$ on the right-hand side of (\ref{eq_mse}) was evaluated numerically. 
We confirmed that the spectral radius $\rho( \bm Q^{(T)})$ was upper bounded by (\ref{eq_mse}) with the MSE loss $L(\bm {x}^{(T)})$.

\section{Permutation search for incremental training emulation}\label{app_c}

The purpose of this appendix is to describe how to emulate incremental training of DUGD using Chebyshev steps to reproduce the zig-zag shape of the learned step sizes.

\begin{figure}[t]\label{alg}
 \removelatexerror
\begin{algorithm}[H]
   \caption{Emulation of incremental training using Chebyshev steps}
   \label{alg_0}
\begin{algorithmic}[1]
   \State {\bfseries Input:} maximum eigenvalue $\lambda_n$, minimum eigenvalue $\lambda_1$, number of iterations $T$, initial value $u\in\mathbb{R}$
   \State Initialize $\bm {c} := 2/(\lambda_1+\lambda_n)$
   \For{$t=2$ {\bfseries to} $T$}
   \State Set $v$ to a sufficiently large number	   
   \State $\bm {d} := ( \bm c, u) $ (concatenation of $\bm  c\in\mathbb{R}^{t-1}$ and $u$)	
   \State Define $\bm c$ as Chebyshev steps of length $t$ with $\lambda_1$ and $\lambda_n$, i.e.,  
   $${\bm c := \left( \left[\frac{\lambda_n+\lambda_1}{2}+\frac{\lambda_n-\lambda_1}{2}\cos\left( \frac{2t+1}{2T}\pi \right)\right]^{-1} \right)_{t=0}^{T-1}}$$
   \For{$\pi$ {\bfseries to all possible permutations} $\Pi(t)$}
   \State Define $\bm P_{\pi}$ as the permutation matrix of $\pi$
   \If{$v>\|\bm {d}-\bm {P}_{\pi}\bm {c}\|_2$}
   \State $v:=\|{\bm d}-\bm  {P}_{\pi}{\bm c}\|_2$, $\bm  P:= \bm P_{\pi}$
   \EndIf
   \EndFor
   \State $ \bm c:= \bm P\bm  c$
   \EndFor
   \State {\bfseries Return:} $\bm c$

\end{algorithmic}
\end{algorithm}
\end{figure}

We consider a training process of DUGD which minimizes the spectral radius $\rho( \bm Q^{(T)})$ instead of the MSE loss function. 
Although it seems practically difficult to solve, we assume that we obtain the Chebyshev steps as an approximate solution.
The problem is which order of the Chebyshev steps is chosen at each generation.
We thus determine an order of the Chebyshev steps that minimizes the ``distance'' from a given initial point to the point whose elements are permuted Chebyshev steps. 
As a measure of distance, we use a simple Euclidean norm
instead of an actual distance defined by $\rho(\bm Q^{(T)})$
 because the latter is difficult to calculate.
The details of the algorithm are shown in Alg.~\ref{alg_0}. 
To emulate incremental training, the length of the Chebyshev steps is gradually increased. As an initial point ($\gamma_1,\dots,\gamma_{t+1}$) of length $t+1$, $\gamma_1, \dots, \gamma_t$ are set to optimally permuted Chebyshev steps in the last generation, and $\gamma_{t+1}$ is set to a given initial value $u$.
Then, an optimal permutation of the Chebyshev steps of length $t+1$ is searched so that its distance from the initial point takes the minimum value. The point is used as an initial point of the next generations. 
As shown in Fig.~\ref{fig_step2}, this successfully reproduces the zig-zag shape of the learned step-size parameters that depend on an initial value of $\gamma_t$.

\section{Proof of Theorem~\ref{convergence_rate}} \label{app_thm3}
 By expanding the SOR update function $S^{(k)}(\bm x)$ around the fixed point $\bm x^*$,
we have 
\begin{equation}
	S^{(k)}(\bm x) = S^{(k)}(\bm x^*) +  \left(\bm I_n - \omega_{k}^{ch} \bm B \right) (\bm x - \bm x^*) 
	+ o(\|\bm x - \bm x^*\|_2),
\end{equation}
where the residual term $o(\|\bm x - \bm x^*\|_2)$ satisfies 
\begin{equation}
	\lim_{\|\bm x - \bm x^*\|_2 \rightarrow 0}\frac{o(\|\bm x - \bm x^*\|_2)}{\|\bm x - \bm x^*\|_2} = \bm 0.
\end{equation}
Substituting $\bm x^{(k)}$ into $\bm x$, the above equality can be rewritten as follows:
\begin{equation}
	\bm x^{(k+1)} - \bm x^* =  \left(\bm I_n - \omega_{k}^{ch} \bm B \right) (\bm x^{(k)} - \bm x^*) + o(\|\bm x^{(k)} - \bm x^*\|_2).
\end{equation}
By recursive applications of the above equation, we eventually get 
\begin{equation}\label{req}
	\bm x^{((\ell+1)T)} - \bm x^* =
\left(\prod_{k = 0}^{T-1} \left(\bm I_n - \omega_{k}^{ch} \bm B \right) \right) (\bm x^{(\ell T)} - \bm x^*) + 
\bm \epsilon,
\end{equation}
where $\bm \epsilon$ is a residual vector satisfying 
\begin{equation}
	\lim_{\ell \rightarrow \infty}\frac{\bm \epsilon}{\|\bm x^{(\ell T)} - \bm x^*\|_2}  = \bm 0.
\end{equation}

By taking norms of (\ref{req}), we obtain
\begin{eqnarray} \nonumber
&& \hspace{-1.3cm}\|\bm x^{((\ell + 1)T)} - \bm x^*\|_2  \\ \nonumber
	&=&
\left\|\left(\prod_{k = 0}^{T-1} \left(\bm I_n - \omega_{k}^{ch} \bm B \right)  \right) (\bm x^{(\ell T)} - \bm x^*) + \bm \epsilon \right\|_2 \\  \nonumber
&\le&
\left\|\left(\prod_{k = 0}^{T-1} \left(\bm I_n - \omega_{k}^{ch} \bm B \right)  \right) (\bm x^{(\ell T)} - \bm x^*)\right\|_2 + \left\|\bm \epsilon \right\|_2 \\ \nonumber 
&\le& 
\left\|\prod_{k = 0}^{T-1} \left(\bm I_n - \omega_{k}^{ch} \bm B \right)  \right\| 
\left\|\bm x^{(\ell T)} - \bm x^*\right\|_2 + \left\|\bm \epsilon \right\|_2 \\ \label{long_ineq}
&=& 
\rho \left(\prod_{k = 0}^{T-1} \left(\bm I_n - \omega_{k}^{ch} \bm B \right)  \right)
\left\|\bm x^{(\ell T)} - \bm x^*\right\|_2 + \left\|\bm \epsilon \right\|_2.
\end{eqnarray}
The first inequality is due to the triangle inequality for any norm and the second inequality is because of 
the sub-additivity of the spectral norm and Euclidean norm. The last equality is due to {$\rho \left(\prod_{k = 0}^{T-1} \left(\bm I_n - \omega_{k}^{ch} \bm B \right)  \right)=\|\prod_{k = 0}^{T-1} \left(\bm I_n - \omega_{k}^{ch} \bm B \right) \|_2 $
because $\bm I_n - \omega_{k}^{ch} \bm B$ is a Hermitian matrix.}
By dividing both sides by $\left\|\bm x^{(\ell T)} - \bm x^*\right\|_2$, we immediately have
\begin{equation}
	\lim_{\ell \rightarrow \infty}\frac{\|\bm x^{((\ell+1)T)} - \bm x^*\|_2}{\|\bm x^{(\ell T)} - \bm x^*\|_2}	
	\le  \rho \left(\prod_{k = 0}^{T-1} \left(\bm I_n - \omega_{k}^{ch} \bm B \right)  \right).
\end{equation}
Applying Lem.~\ref{lem_qt}, the claim of the theorem is given.
 \hfill \IEEEQEDclosed
 
\section{Proof of Theorem~\ref{real_th}}\label{app_thm4}

From the assumption $g'(x) \ge 0$, $\bm Q^*$ is a diagonal matrix with non-negative diagonal elements.
Because $\bm A \in \mathbb{R}^{n \times n}$ is a full-rank matrix, the rank of $\bm Q^* \bm A$ coincides with
the number of non-zero diagonal elements in $\bm Q^*$ which is denoted by $r$.
The Darazin-Hynsworth theorem \cite{Drazin62} presents that the necessary and sufficient condition 
for $\bm X \in \mathbb{C}^{n \times n}$ having $m (\le n)$ linearly independent eigenvectors 
and corresponding $m$ real eigenvalues is that there exists a symmetric semi-positive definite matrix $\bm S$ 
with rank $m$ satisfying 
$
	\bm X \bm S = \bm S \bm X^H.	
$
The matrix $\bm X^H$ represents the Hermitian transpose of $\bm X$.
As we saw, $\bm Q^*$ is a symmetric semi-positive definite matrix with rank $r$.
Multiplying $\bm Q^*$ with $\bm Q^* \bm A$ from the right, we have the equality
\begin{equation}
	(\bm Q^* \bm A) \bm Q^* = \bm Q^* (\bm A \bm Q^*) =  \bm Q^* (\bm Q^{*T} \bm A^T)^T = \bm Q^* (\bm Q^* \bm A)^T,
\end{equation}
which satisfies the necessary and sufficient condition of the Darazin-Hynsworth theorem~\cite{Drazin62}.
This implies that $\bm Q^* \bm A$ has $r$ real eigenvalues. 
\hfill \IEEEQEDclosed

\end{document}